\documentclass[letterpaper]{article}
\usepackage{isea}
\usepackage[pdftex]{graphicx}
\usepackage{times}
\usepackage{helvet}
\usepackage{courier}
\usepackage[hyphens]{url}
\usepackage{wrapfig,booktabs}
\usepackage[colorlinks]{hyperref}
\usepackage{url}
\usepackage{xcolor}
\usepackage{amsmath,amssymb,amsthm}
\usepackage[ruled]{algorithm2e}
\usepackage{wrapfig}
\usepackage{subcaption}
\usepackage{graphicx}
\usepackage[export]{adjustbox}
\usepackage{float}
\usepackage{mathtools, nccmath}
\hypersetup{breaklinks=true}
\usepackage[numbers]{natbib}

\title{Developing Creative AI to Generate Sculptural Objects}
\author{Songwei Ge\thanks{Equal contribution}, Austin Dill$^{\star}$, Eunsu Kang$^{\star}$, Chun-Liang Li$^{\star}$, Lingyao Zhang\\
{\bf \Large Manzil Zaheer, Barnabas Poczos} \\
Carnegie Mellon University\\
Pittsburgh, PA, United States \\
\{chunlial, eunsuk, songweig, lingyaoz, abdill, manzil, bapoczos\}@andrew.cmu.edu\\
\newline
\newline
}
\begin{document} 
\maketitle
\begin{abstract}
We explore the intersection of human and machine creativity by generating sculptural objects through machine learning. This research raises questions about both the technical details of automatic art generation and the interaction between AI and people, as both artists and the audience of art. We introduce two algorithms for generating 3D point clouds and then discuss their actualization as sculpture and incorporation into a holistic art installation. Specifically, the Amalgamated DeepDream (ADD) algorithm solves the sparsity problem caused by the naive DeepDream-inspired approach and generates creative and printable point clouds. The Partitioned DeepDream (PDD) algorithm further allows us to explore more diverse 3D object creation by combining point cloud clustering algorithms and ADD.

\end{abstract}


Partitioned DeepDream, Amalgamated DeepDream, 3D, Point Cloud, Sculpture, Art, Interactive Installation, Creative AI, Machine Learning

\section{Introduction}

Will Artificial Intelligence (AI) replace human artists or will it show us a new perspective into creativity? Our team of artists and AI researchers explore artistic expression using Machine Learning (ML) and design creative ML algorithms to be possible co-creators for human artists. 

In terms of AI-generated and AI-enabled visual artwork, there has been a good amount of exploration done over the past three years in the 2D image area traditionally belonging to the realm of painting. Meanwhile, there has been very little exploration in the area of 3D objects, which traditionally would belong to the realm of sculpture and could be easily extended into the area of art installations. The creative generation of 3D object research by Lehman et al. successfully generated ``evolved'' forms, however, the final form was not far from the original form and could be said to merely mimic the original~\citep{lehman2016creative}. Another relevant study, Neural 3D Mesh Renderer~\citep{kato2018neural} focuses on adjusting the rendered mesh based on DeepDream~\citep{deepdream} textures. In the art field, artist Egor Kraft's Content Aware Studies project~\cite{contentaware} explores the possibilities of AI to reconstruct lost parts of antique Greek and Roman sculptures. Nima et al.~\cite{nima2018vox2net} introduced Vox2Net adapted from pix2pix GAN which can extract the abstract shape of the input sculpture. DeepCloud~\cite{DeepCloud} exploited an autoencoder to generate point clouds, and constructed a web interface with analog input devices for users to interact with the generation process. Similar to the work of Lehman et al~\citep{lehman2016creative}, the results generated by DeepCloud are limited to previously seen objects.

\begin{figure}[!t]
    \begin{subfigure}{.47\linewidth}
        \centering
        \includegraphics[height=1.1\linewidth]{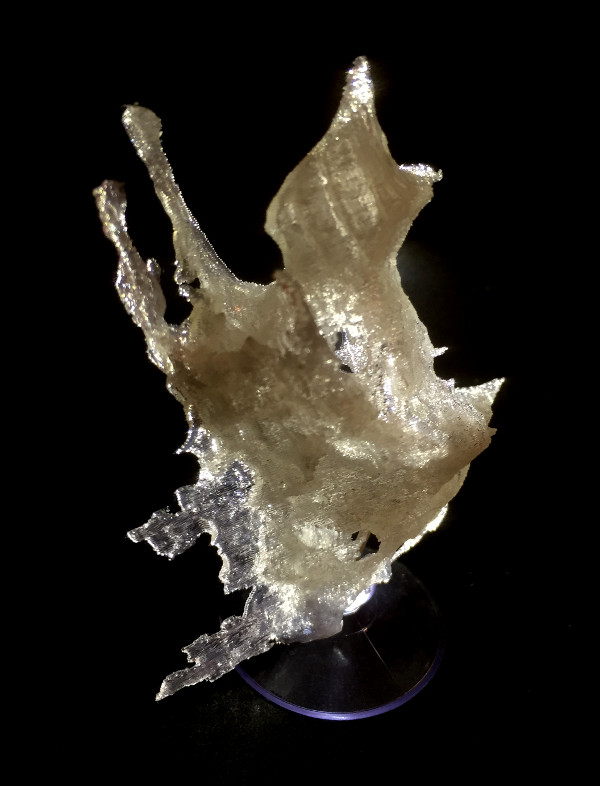}
    \end{subfigure}
    \hfill
	\begin{subfigure}{.47\linewidth}
	    \centering
        \includegraphics[height=1.1\linewidth]{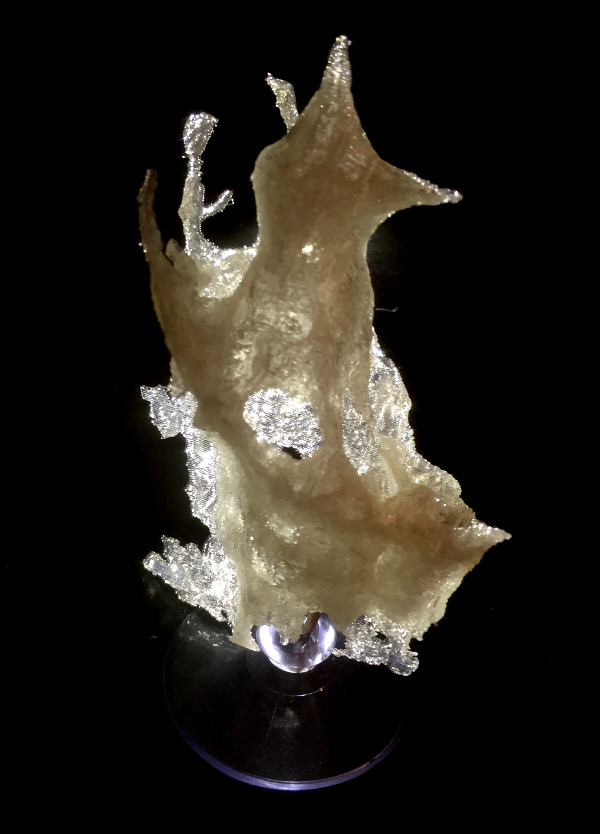}
	\end{subfigure}
	\caption{Sculpture generated by creative AI, PDD.}
	\label{fig:print_pdd}
\end{figure}

In this paper, we introduce two ML algorithms to create original sculptural objects. As opposed to previous methods, our results do not mimic any given forms and are original and creative enough not to be categorized into the dataset categories. Our method extends an idea inspired by DeepDream for 2D images to 3D point clouds. However, simply translating the method generates low-quality objects with local and global sparsity issues that undermine the integrity of reconstruction by 3D printing. Instead we propose Amalgamated DeepDream (ADD) which utilizes union operations during the process to generate both creative and realizable objects. Furthermore, we also designed another ML generation algorithm, Partitioned DeepDream (PDD), to create more diverse objects, which allows multiple transformations to happen on a single object. With the aid of mesh generation software and 3D printing technology, the generated objects can be physically realized. In our latest artwork, it has been incorporated into an interactive art installation.

\section{Creative AI: ADD and PDD}

Artistic images created by AI algorithms have drawn great attention from artists. AI algorithms are also attractive for their ability to generate 3D objects which can assist the process of sculpture creation. In graphics, there are many methods to represent a 3D object such as mesh, voxel and point cloud. In our paper, we focus on point cloud data which are obtained from two large-scale 3D CAD model datasets: ModelNet40~\cite{wu20153d} and ShapeNet~\cite{chang2015shapenet}. These datasets are preprocessed by uniformly sampling from the surface of the CAD models to attain the desired number of points. Point cloud is a compact way to represent 3D object using a set of points on the external surfaces and their coordinates such as those shown in Figures \ref{fig:ModelNet40} and \ref{fig:ShapeNet}. We introduce two algorithms, ADD and PDD, which are inspired by 2D DeepDream to generate creative and realizable point clouds. In this section, we will briefly discuss the existing 3D object generation methods, and then elaborate the ADD and PDD algorithms and present some of the generated results.

\begin{minipage}[T]{\linewidth}
      \vspace{-1.4em}
      \begin{minipage}{0.48\linewidth}
          \begin{figure}[H]
              \begin{subfigure}[b]{.48\linewidth}
              \centering
              \includegraphics[trim={100 200 100 50},clip,width=\linewidth]{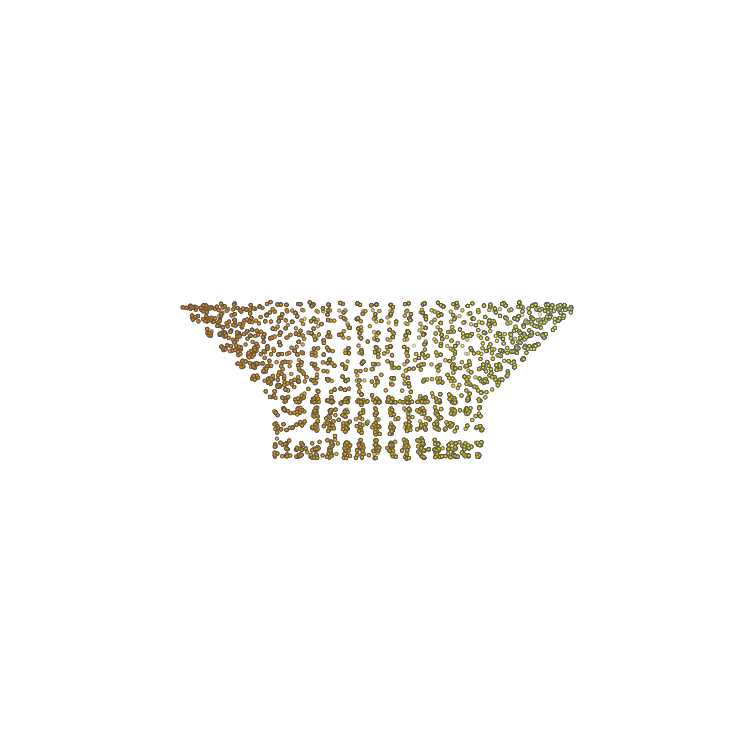}
              \end{subfigure}
              \begin{subfigure}[b]{.48\linewidth}
              \centering
              \includegraphics[trim={100 200 100 50},clip,width=\linewidth]{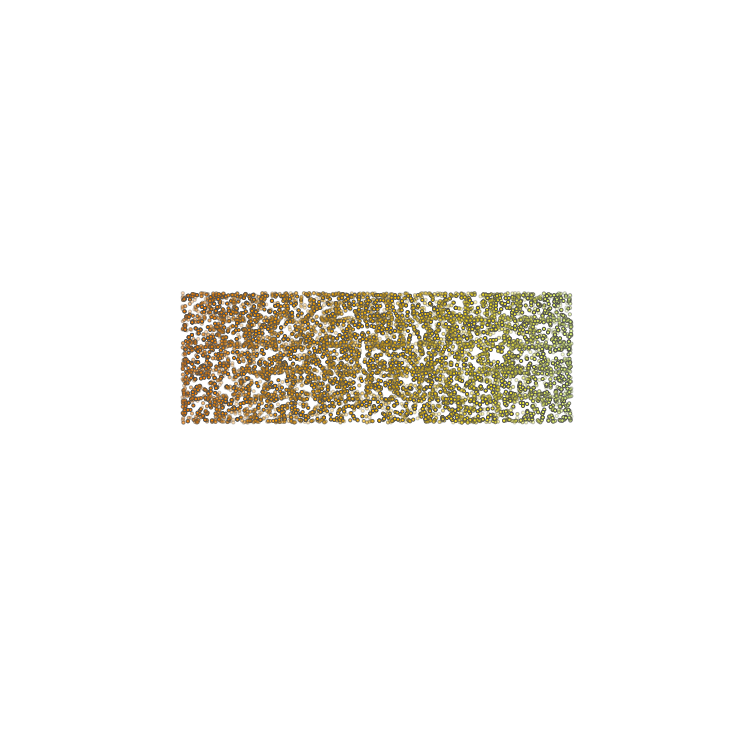}
              \end{subfigure}
              \caption{ModelNet40~\cite{wu20153d}}
              \label{fig:ModelNet40}
          \end{figure}
      \end{minipage}
      \hfill
      \begin{minipage}{0.48\linewidth}
          \begin{figure}[H]
              \centering
              \begin{subfigure}[b]{.48\linewidth}
              \centering
              \includegraphics[trim={20 150 180 100},clip,width=\linewidth]{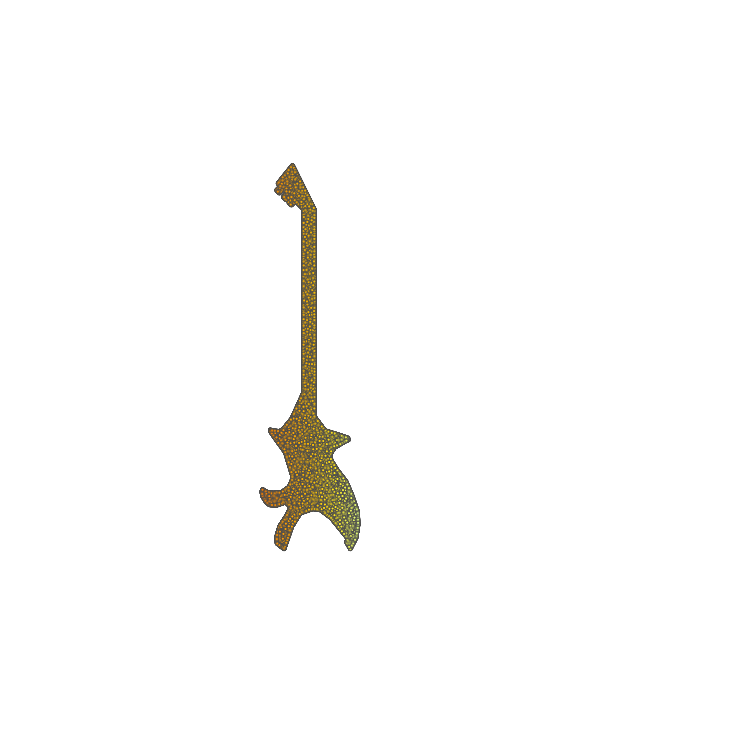}
              \end{subfigure}
              \begin{subfigure}[b]{.48\linewidth}
              \centering
              \includegraphics[trim={100 150 100 100},clip,width=\linewidth]{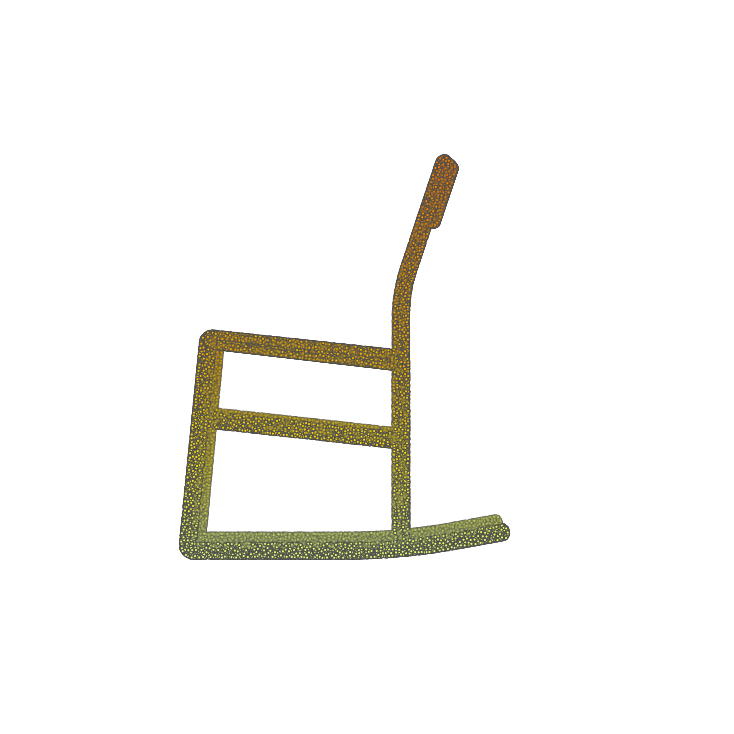}
              \end{subfigure}
              \caption{ShapeNet~\cite{chang2015shapenet}}
              \label{fig:ShapeNet}
          \end{figure}
      \end{minipage}
      \hfill
\end{minipage}


\subsection{Learning to Generate Creative Point Clouds}

\begin{figure}[!h]
\centering
\adjincludegraphics[width=0.9\linewidth]{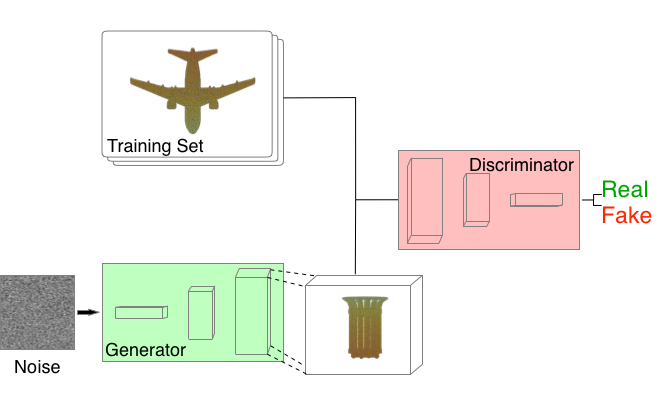}
\caption{An example of generative model.}
\label{fig:imagegan}
\end{figure}

Using deep learning to generate new objects has been studied in different data types, such as music~\citep{van2016wavenet}, images~\citep{goodfellow2014generative}, 3D voxels~\citep{wu2016learning} and point clouds~\citep{achlioptas2017learning, li2018point}. An example of a simple generative model is shown in Figure \ref{fig:imagegan}.

Here a low-dimensional latent space $h$ of the original object space $x$ is learned in accordance with the underlying probability distribution. 
\begin{equation}
    p(x) = \int_h p(h)p(x|h) dh
\end{equation}
A discriminator is usually used to help distinguish the generated object $\hat{x}$  and a real object $x$. Once the discriminator is fooled, the generator can create objects that look very similar to the original ones \cite{goodfellow2014generative}.
However, these generative models only learn to generate examples from the ``same'' distribution of the given training data, instead of learning to generate ``creative'' objects~\citep{elgammal2017can,nima2018vox2net}.  


\begin{figure}
    \centering
    \includegraphics[width=\linewidth]{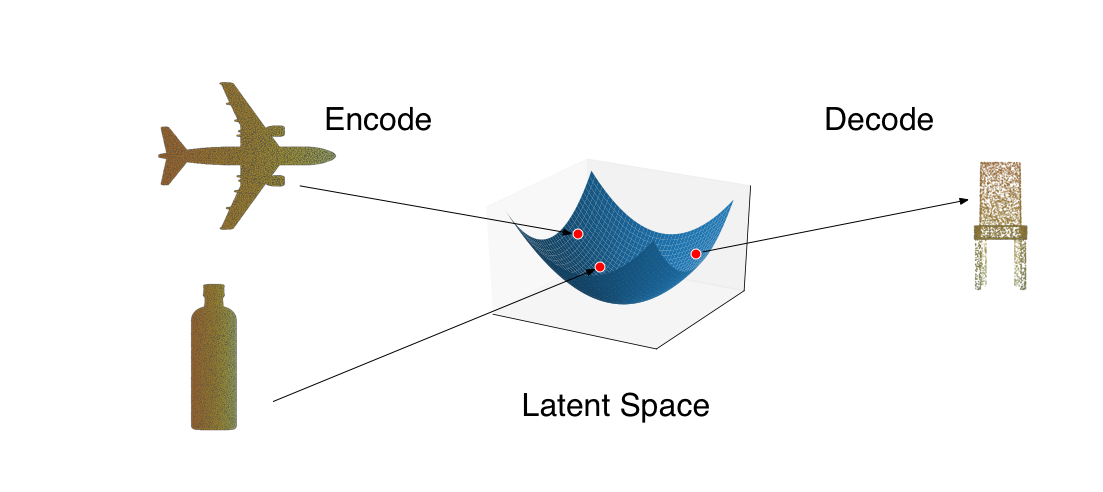}
    \caption{Encoding and decoding in latent space}
    \label{fig:latent_space}
\end{figure}

\begin{figure}[!h]
\centering
\begin{subfigure}{.3\linewidth}
\centering
\adjincludegraphics[trim={{0.3\width} {0.1\height} {0.3\width} 0}, clip, width=0.33\linewidth]{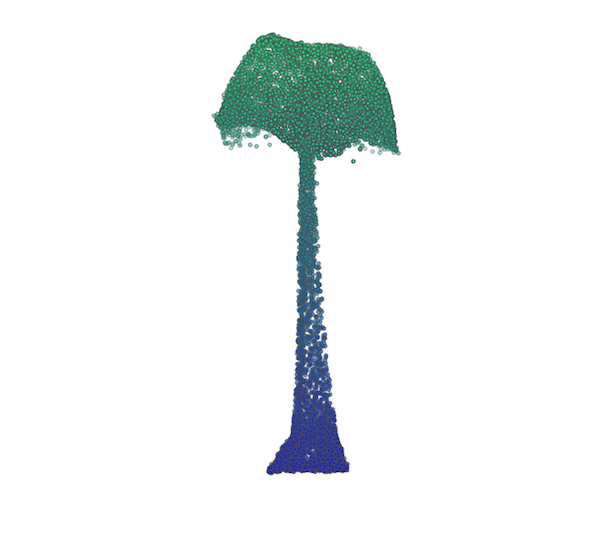}
\caption{Sampling}
\end{subfigure}
\hspace*{-0.2cm}
\begin{subfigure}{.3\linewidth}
\centering
\adjincludegraphics[trim={{0.2\width} {0.2\height} {0.2\width} {0.1\height}}, clip, width=0.6\linewidth]{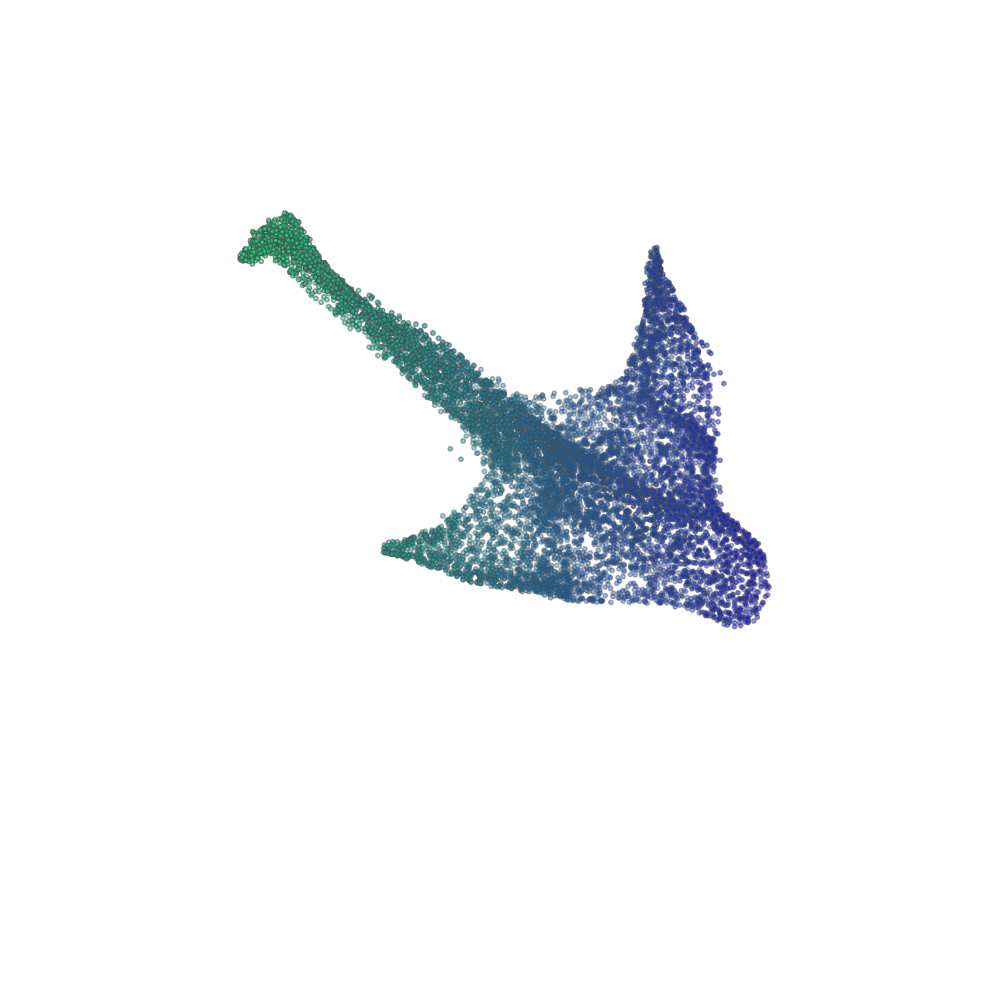}
\caption{Sampling}
\end{subfigure}
\begin{subfigure}{.33\linewidth}
\centering
\adjincludegraphics[trim={{0.2\width} {0.1\height} {0.2\width} {0.1\height}}, clip, width=0.5\linewidth]{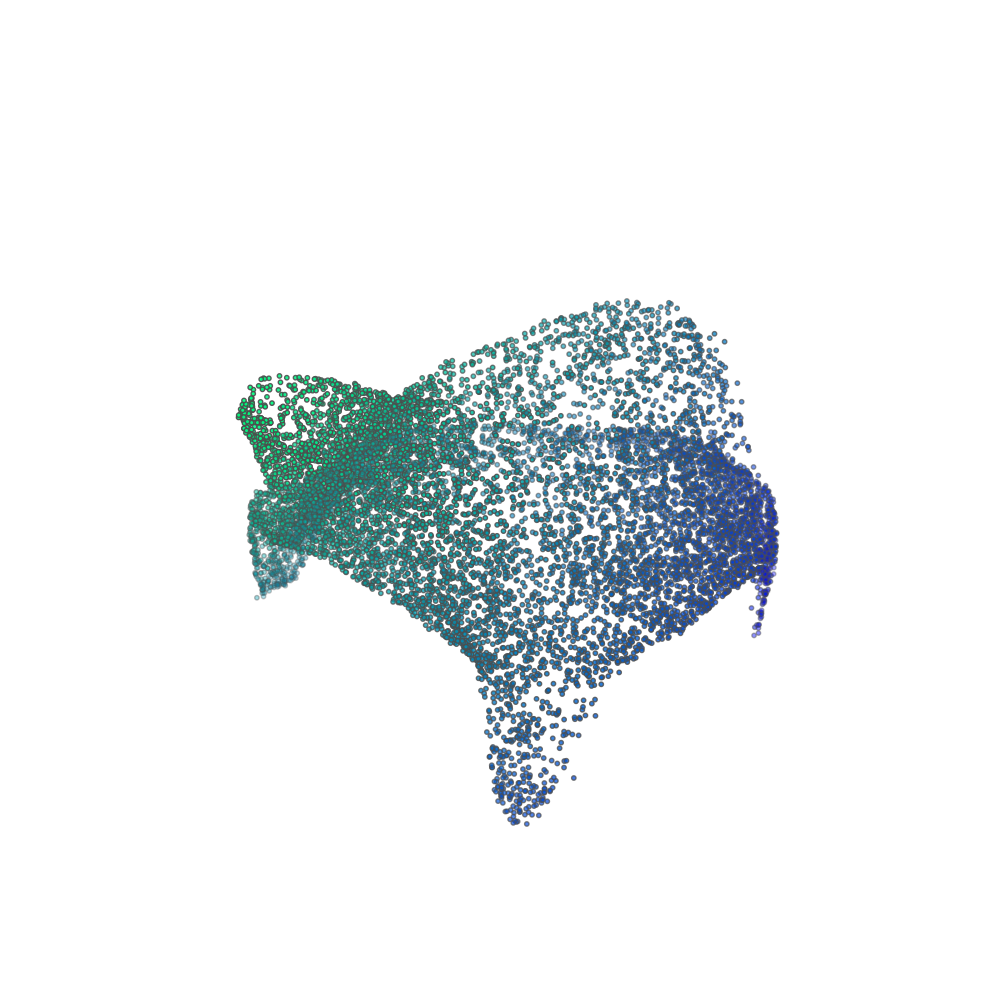}
\caption{Sampling}
\end{subfigure}

\begin{subfigure}{.3\linewidth}
\centering
\adjincludegraphics[trim={{0.15\width} {0.1\height} {0.15\width} {0.1\height}}, clip, width=0.55\linewidth]{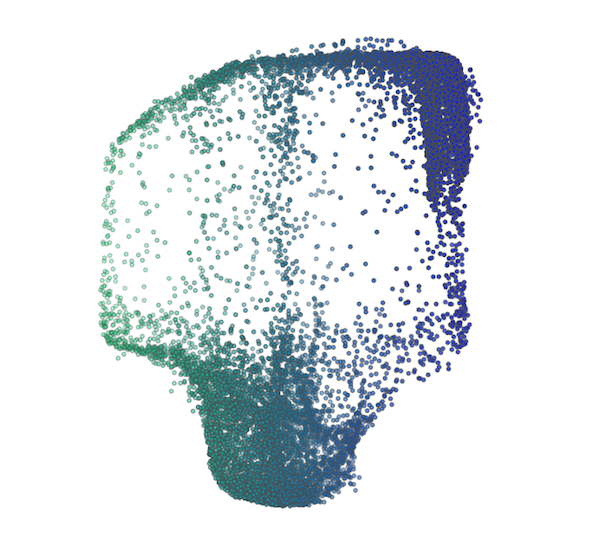}
\caption{Interpolation}
\end{subfigure}
\begin{subfigure}{.3\linewidth}
\centering
\adjincludegraphics[trim={{0.25\width} {0.2\height} {0.2\width} {0.15\height}}, clip, width=0.43\linewidth]{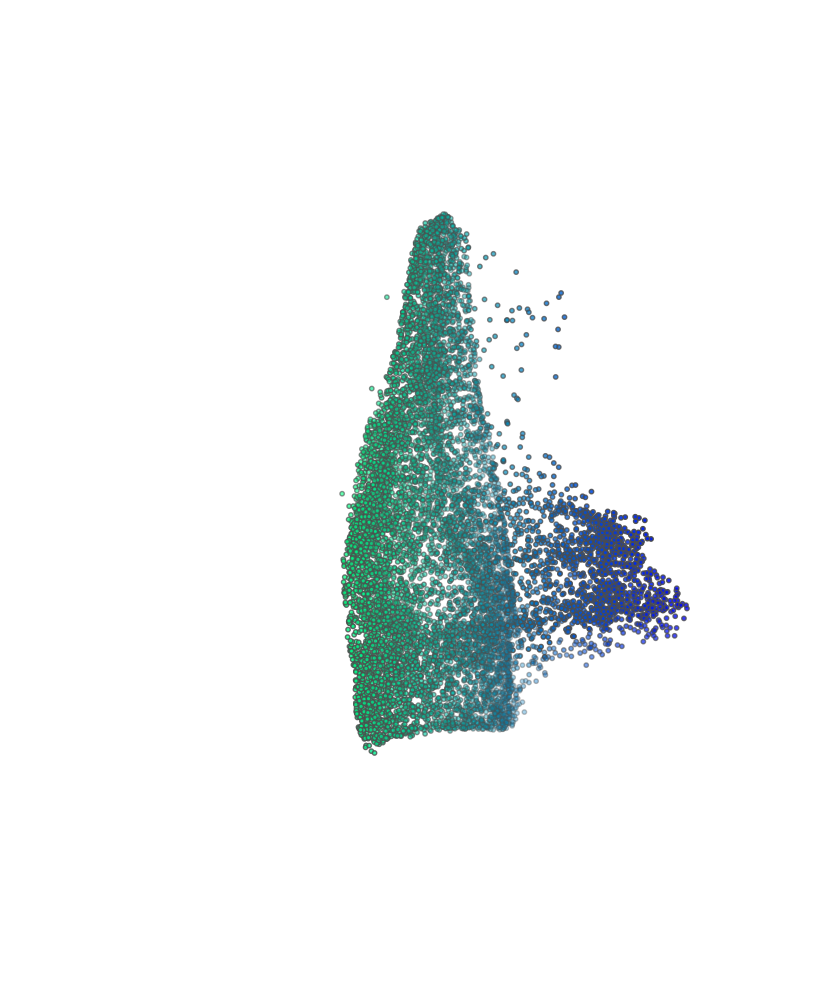}
\caption{Interpolation}
\end{subfigure}
\begin{subfigure}{.33\linewidth}
\centering
\adjincludegraphics[trim={{0.2\width} {0.25\height} {0.2\width} {0.15\height}}, clip, width=0.53\linewidth]{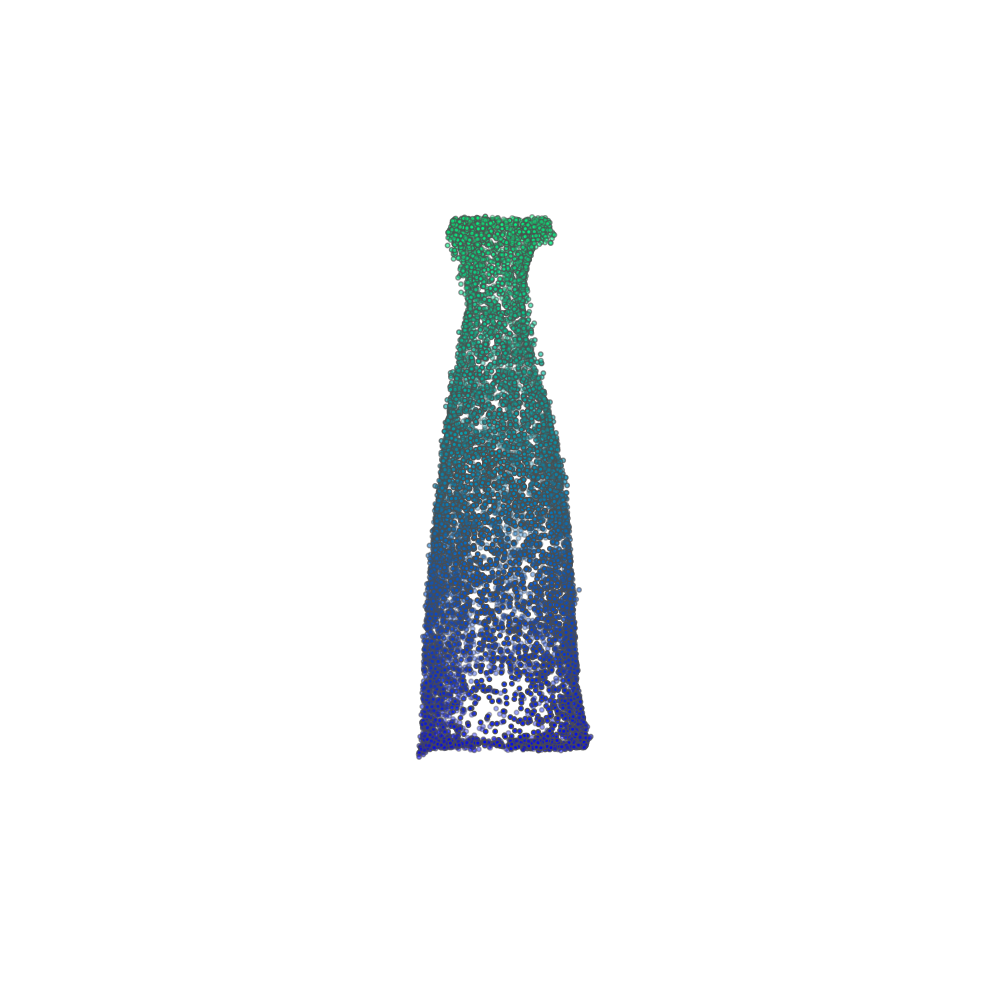}
\caption{Interpolation}
\end{subfigure}
\caption{Results of~\citet{li2018point}.}
\label{fig:pcgan}
\end{figure}

One alternative is to decode the convex combination $\hat{h}$ of latent codes $h_1, h_2$ of two objects in an autoencoder to get final object $\hat{X}$. Specifically, an autoencoder is a kind of feedforward neural network used for dimensionality reduction whose hidden units can be viewed as latent codes which capture the most important aspects of the object~\cite{robert2014machine}. A representation of the process of encoding to and decoding from latent space can be seen in Figure \ref{fig:latent_space}.

\begin{align*}
    \label{algo: interpo}
    \hat{h} &= t \cdot h_1 + (1-t) \cdot h_2  \\
    \hat{x} &= \texttt{Generator}(\hat{h}),\\
\end{align*}
where $ 0 \leq t \leq 1 $. Empirical evidence shows that decoding mixed codes usually produces semantically meaningful objects with features from the corresponding objects. This approach has also been applied to image creation~\cite{carter2017using}. One could adopt  encoding-decoding algorithms for point clouds~\citep{yang2018foldingnet, groueix2018atlasnet, li2018point} based on the same idea. The sampling and interpolation results based on~\citet{li2018point} are shown in Figure~\ref{fig:pcgan}. 

\paragraph{Inspiration: DeepDream for Point Clouds} 

\begin{figure}
    \centering
    \includegraphics[width=\linewidth]{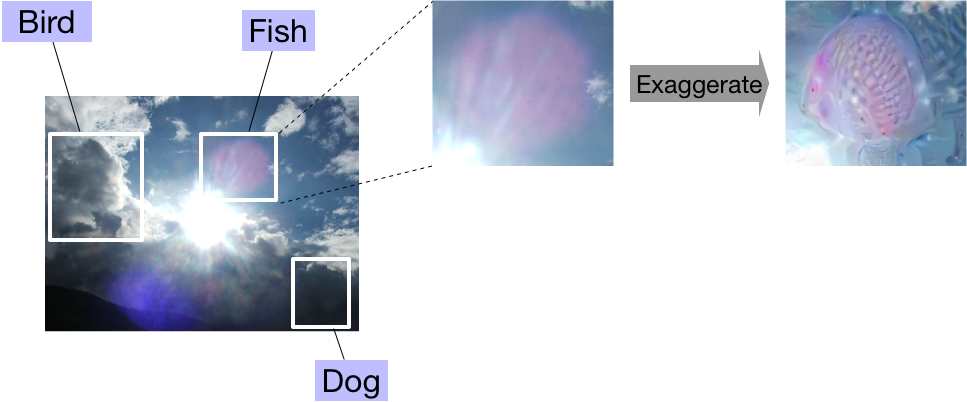}
    \caption{2D DeepDream visualized}
    \label{fig:deepdream_vis}
\end{figure}

In contrast to simple generative models and interpolations, DeepDream leverages trained deep neural networks by enhancing patterns to create psychedelic and surreal images.  For example in Figure~\ref{fig:deepdream_vis}, the neural network detects features in the input images similar to the object classes of bird, fish and dog and then exaggerates these underlying features. Given a trained neural network $f_\theta$ and an input image $x$,  DeepDream aims to modify $x$ to maximize (amplify) $f_\theta(x; a)$, where $a$ is an activation function of $f_\theta$. After this process, $x$ is expected to display some features that are captured by $f_\theta(x; a)$. Algorithmically, DeepDream iteratively modifies $x$ via a gradient update with a certain learning rate $\gamma$.
\begin{equation}
	x_t = x_{t-1} + \gamma \nabla_x f_\theta(x; a). 
	\label{eq:gradient}
\end{equation}

\begin{wrapfigure}[14]{r}{0.249\textwidth}
\centering
    \includegraphics[width=0.8\linewidth]{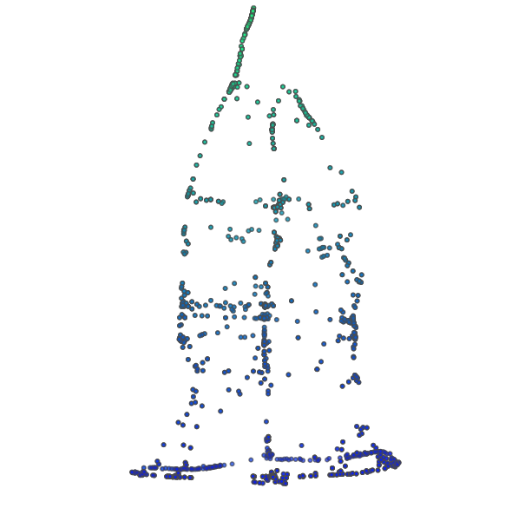}
\caption{Naive DeepDream with undesirable sparse surface.}
\label{fig:dd}
\end{wrapfigure}

An extension of DeepDream is to use a classification network as $f_\theta$ and we replace $a$ with the outputs of the final layer corresponding to certain labels.  This is related to adversarial attack~\citep{szegedy2013intriguing} and unrestricted adversarial examples~\citep{unrestricted_Advex_2018, liu2018adversarial}. 

Unfortunately, directly extending the idea of DeepDream to point clouds with neural networks for sets~\citep{qi2017pointnet, zaheer2017deep} results in undesirable point clouds suffering from both local and global sparsity as shown in Figure~\ref{fig:dd}. To be specific, iteratively applying gradient update without any restrictions creates local holes in the surface of the object. Besides, the number of input points per object is limited by the classification model. Consequently the generated object is not globally dense enough to transform into mesh structure, let alone realize in the physical world.

To avoid local sparsity in the generated point cloud, one compromise is to run the naive 3D point cloud DeepDream with fewer iterations or a smaller learning rate, but it results in limited difference from the input point cloud. Another approach to solving the sparsity problem is conducting post-processing to add points to sparse areas. However, without prior knowledge of the geometry of the objects, making smooth post-processed results is non-trivial~\citep{hoppe1992surface}. For this reason, we simply take inspiration from the gradient update strategy of DeepDream, while developing our novel algorithms. 

\subsection{Amalgamated DeepDream (ADD)}


\begin{figure}
    \begin{subfigure}{.48\linewidth}
        \centering
        \includegraphics[trim={150 200 150 150},clip,width=0.9\linewidth]{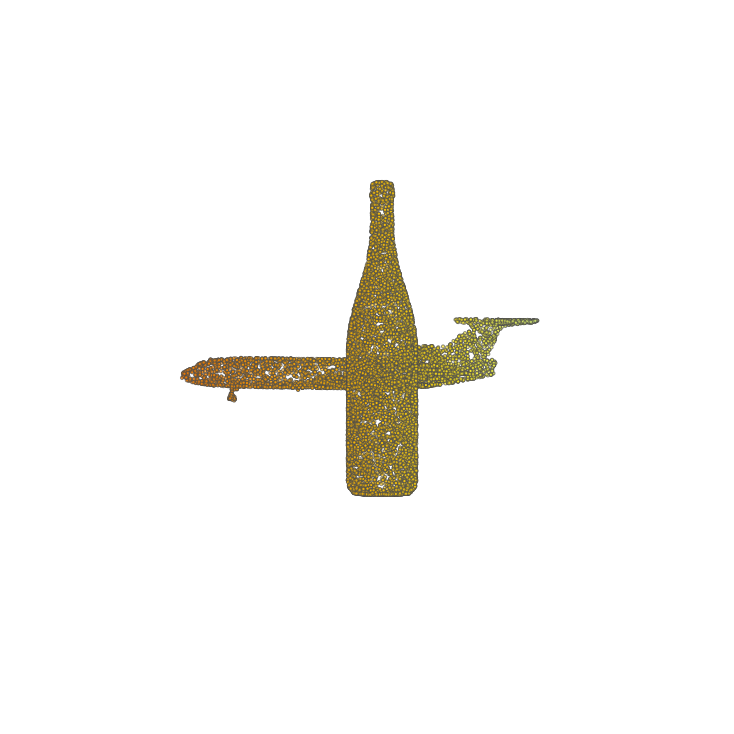}
        \caption{Union of airplane and bottle}
    \end{subfigure}
	\begin{subfigure}{.48\linewidth}
	    \centering
        \includegraphics[trim={150 160 150 190}, clip, width=0.9\linewidth]{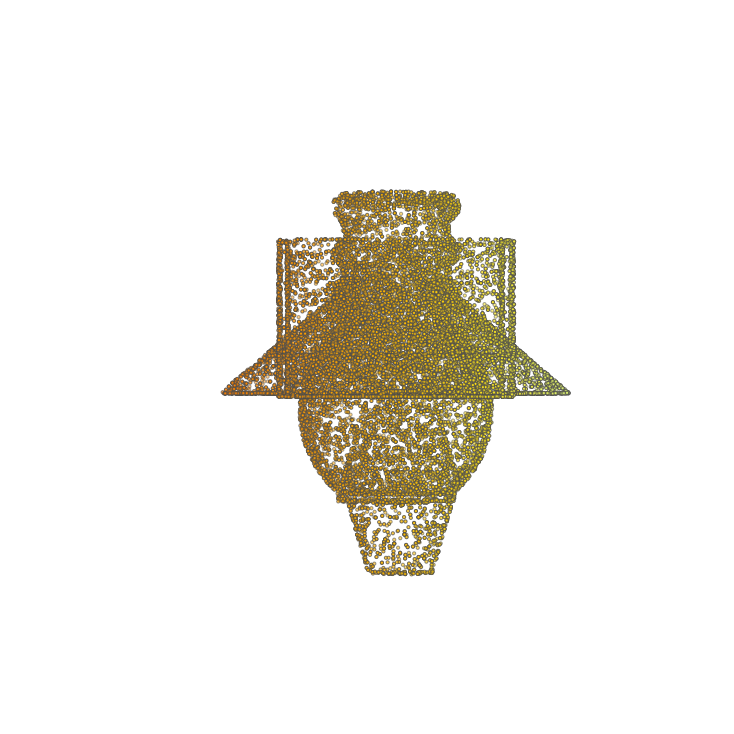}
        \caption{Union of cone, vase and lamp}
	\end{subfigure}
	\caption{Amalgamated input for ADD.}
	\label{fig:union}
\end{figure}

As opposed to images, mixing two point clouds is a surprisingly simple task, which can be done by taking the union of two point clouds because of their permutation invariance property. For example, the union of two or three objects is shown in figure~\ref{fig:union}. This observation inspires us to use a set-union operation, which we call amalgamation, in the creation process to address the drawbacks of naive extension.

\begin{algorithm}
	\caption{Amalgamated DeepDream (ADD)}
  	\SetKwInOut{Input}{input}\SetKwInOut{Output}{output}
	\Input{trained $f_\theta$ and input $X$}
	\Output{generated object $\hat{X}$}
	\For{$x = X_0,\ldots,X_n$}{
	    $x_0 = x$ \\
    	\For{$t=1\dots T$} {
    		$\hat{x} = x_{t-1} + \gamma \nabla_x f_\theta(x_{t-1}; a)$ \\
    		$x_t = \hat{x}$ \\
    		\If{$t$ is divisible by 10}{
    		    $x_t = x_t \cup x$ \\
    		    down-sample $x_t$ to fix the number of points
    		}
    	}
    	$\hat{X} = \hat{X} \cup x_T$
	}
	\label{algo:add}
\end{algorithm}

\begin{figure}[!b]
	\begin{subfigure}[b]{.32\linewidth}
	\centering
	\includegraphics[trim={170 200 210 150},clip,width=\linewidth]{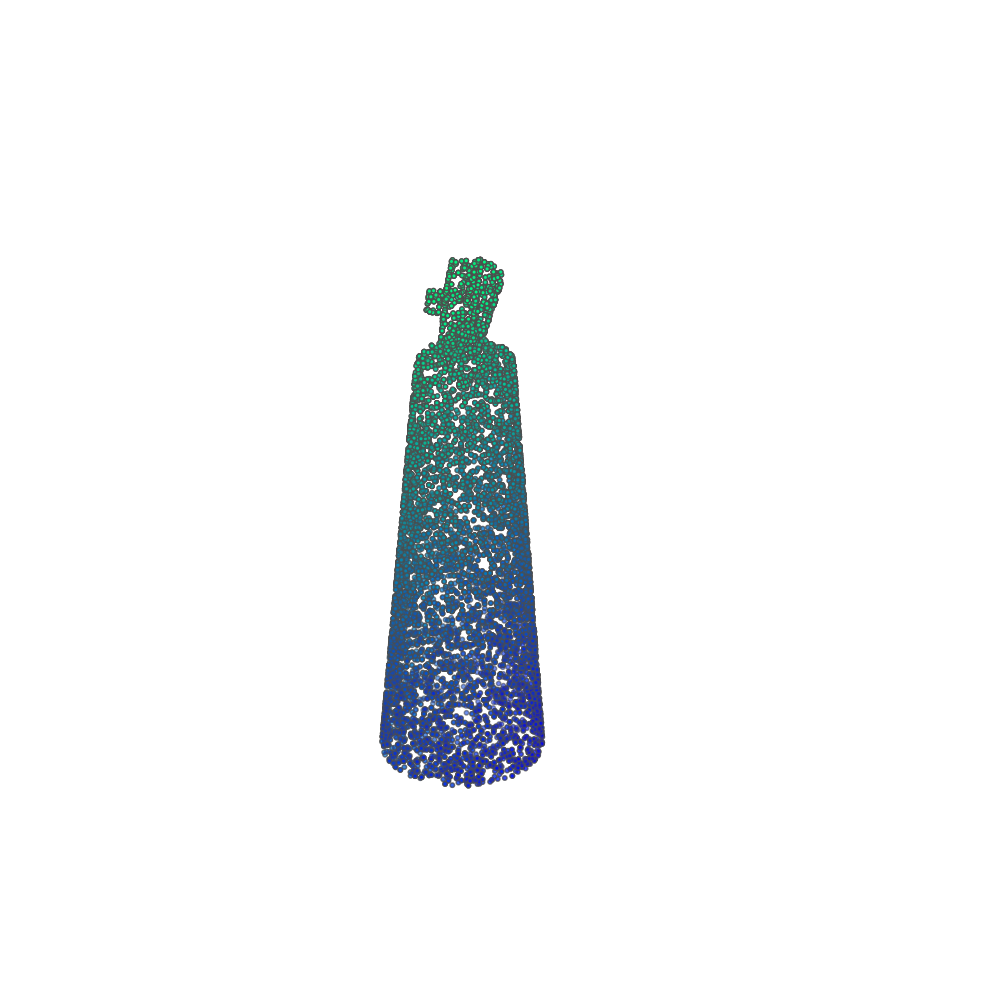}
	\caption{Input}
	\end{subfigure}
	\begin{subfigure}[b]{.32\linewidth}
	\centering
	\includegraphics[trim={170 200 210 150},clip,width=\linewidth]{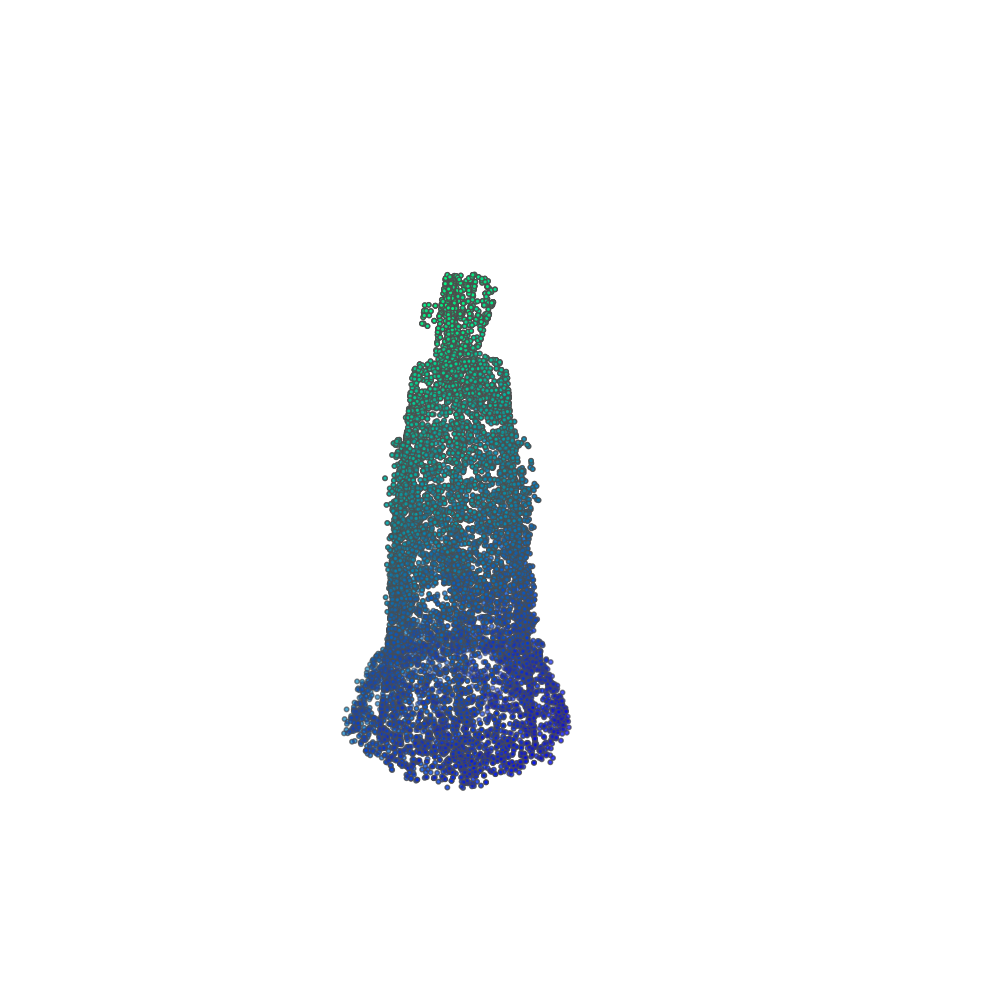}
	\caption{Iteration 10}
	\end{subfigure}
	\begin{subfigure}[b]{.32\linewidth}
	\centering
	\includegraphics[trim={170 200 210 150},clip,width=\linewidth]{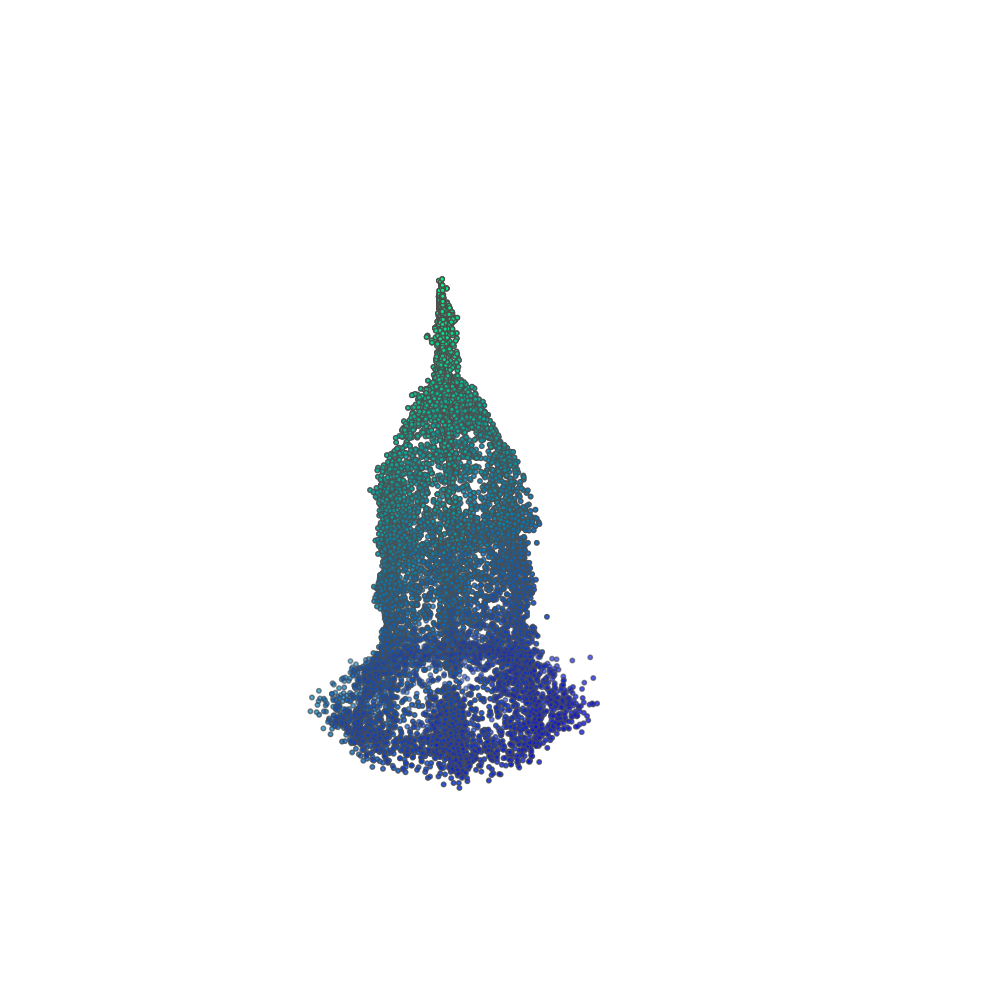}
	\caption{Iteration 100}
	\end{subfigure}
	\caption{Transforming a bottle into a cone via ADD.}
	\label{fig:add}
\end{figure}

For object $X$ with any number of points, we first randomly take it apart into several subsets with the number of points that is required by the classification model, and then modify the input image through gradient updating~\eqref{eq:gradient}. In the end, we amalgamate all the transformed subsets into a final object. By doing this we can generate as dense objects as we want as long as the input object allows. To solve the local sparsity, when running the gradient update~\eqref{eq:gradient} for each subset we amalgamate the transformed point clouds with the input point clouds after every 10 iterations. To avoid exploded number of points, we also down-sample the object after each amalgamation operation. We call the proposed algorithm Amalgamated DeepDream (ADD) as shown in Algorithm~\ref{algo:add}.

\begin{figure}[!t]
    \begin{subfigure}{.48\linewidth}
        \centering
        \includegraphics[trim={150 200 150 150},clip,width=0.9\linewidth]{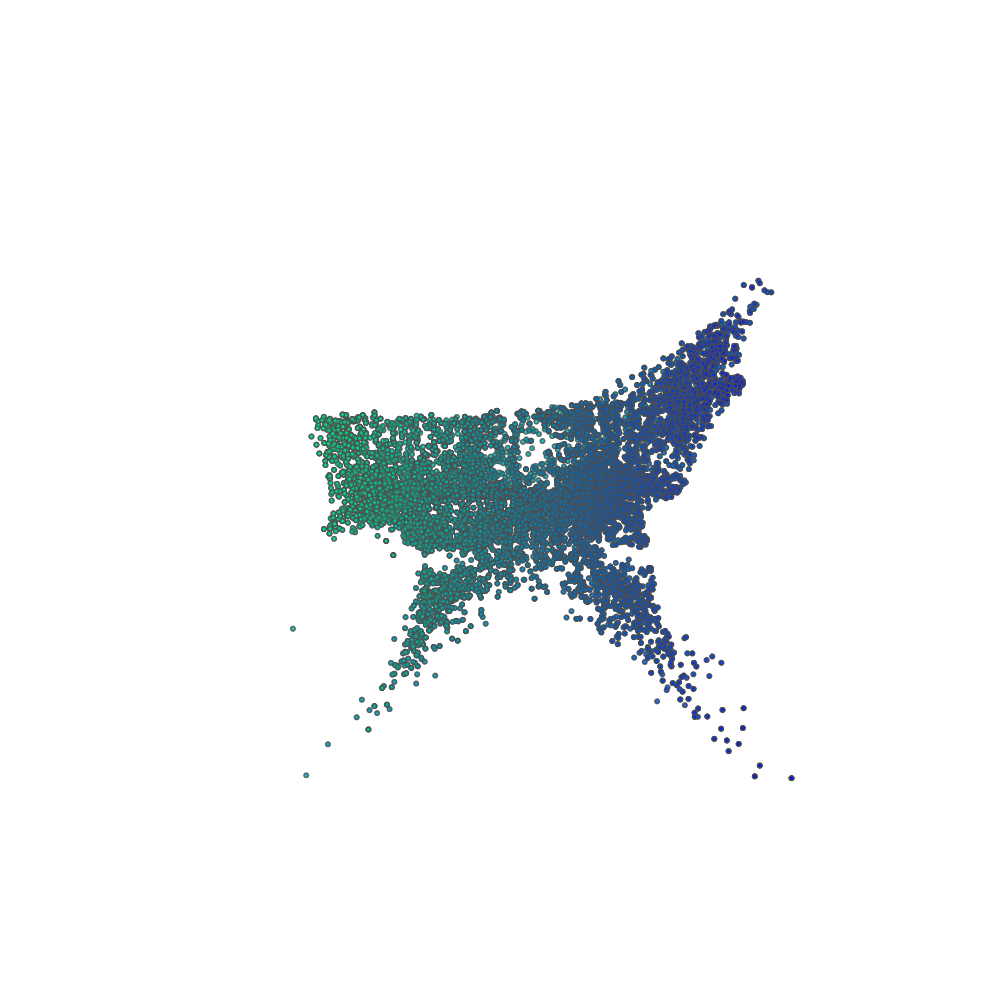}
        \caption{Input: bowl; Target: chair}
    \end{subfigure}
	\begin{subfigure}{.48\linewidth}
	    \centering
        \includegraphics[trim={150 200 150 150},clip,width=0.9\linewidth]{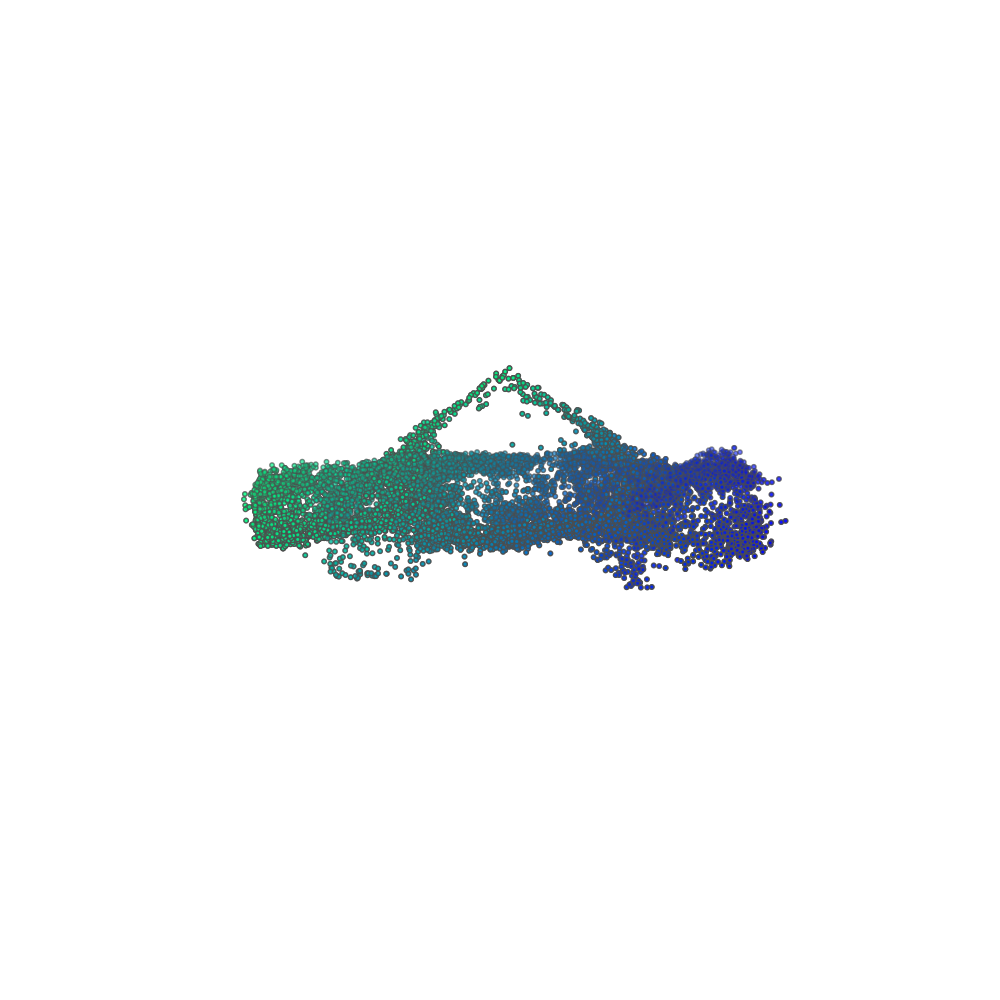}
        \caption{Input: keyboard; Target: car}
	\end{subfigure}
	\caption{ADD with single object input.}
	\label{fig:single}
\end{figure}

\paragraph{Experiments and results} To test our methods, we use DeepSet~\cite{zaheer2017deep} with 1000 points capacity as our basic 3D deep neural network model. We randomly sampled 10000 points from the CAD models and feed them into ADD model for 100 iterations with learning rate set to 1. A running example of ADD targeting the transformation of bottle into cone in ModelNet40~\cite{wu20153d}, which is the same as Figure~\ref{fig:dd}, is shown in Figure~\ref{fig:add}. During the transformation process, ADD enjoys the advantage of deforming objects based on gradient updates with respect to a trained neural network as DeepDream without creating obviously local sparse areas. It can also better preserve the features of input point clouds with the amalgamation operation when we create new features based on DeepDream updates. In addition, the amalgamation of several transformed subsets from original objects allows us to generate denser objects that are realizable in the physical world. More created point clouds from the objects in Figure~\ref{fig:ModelNet40} are shown in Figure~\ref{fig:single}.

\begin{figure}[!b]
    \begin{subfigure}{.48\linewidth}
        \centering
        \includegraphics[trim={150 200 150 150},clip,width=0.9\linewidth]{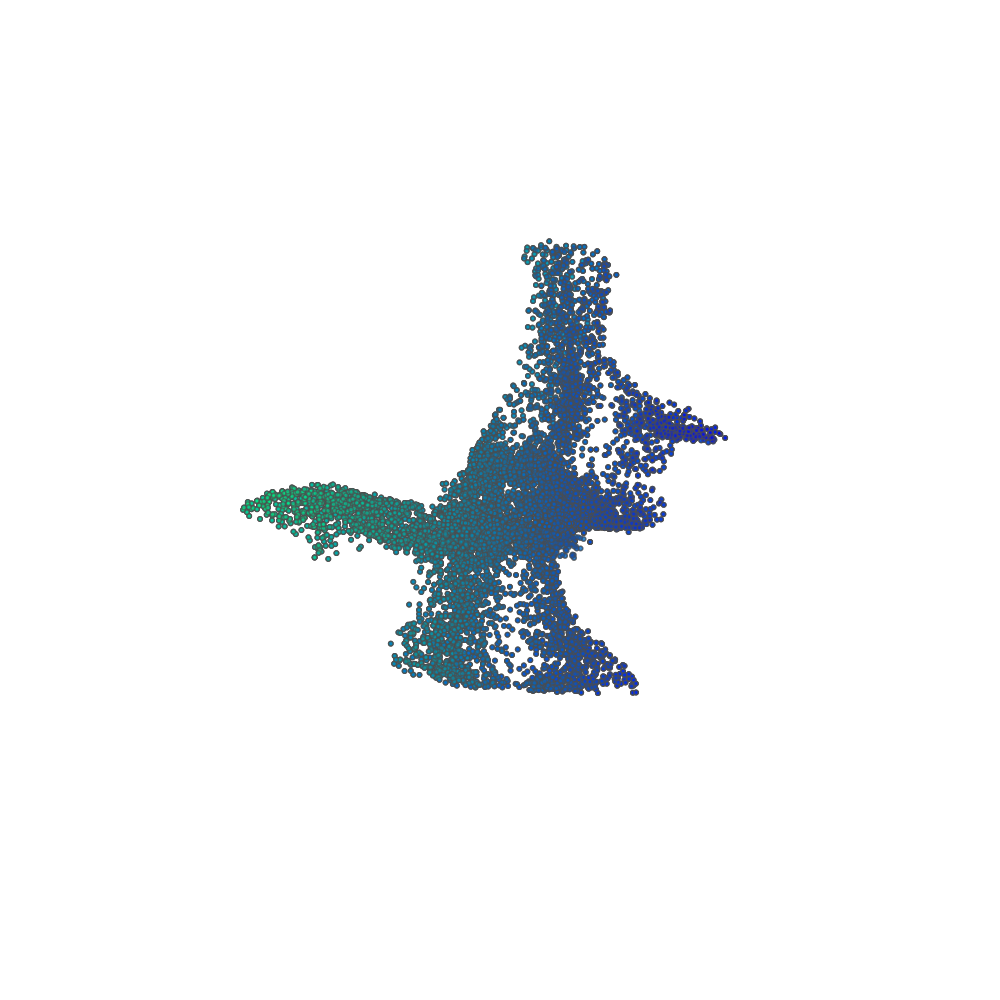}
        \caption{Input: airplane, bottle; \\ Target: toilet}
    \end{subfigure}
	\begin{subfigure}{.48\linewidth}
	    \centering
        \includegraphics[trim={220 180 130 220},clip,width=0.9\linewidth]{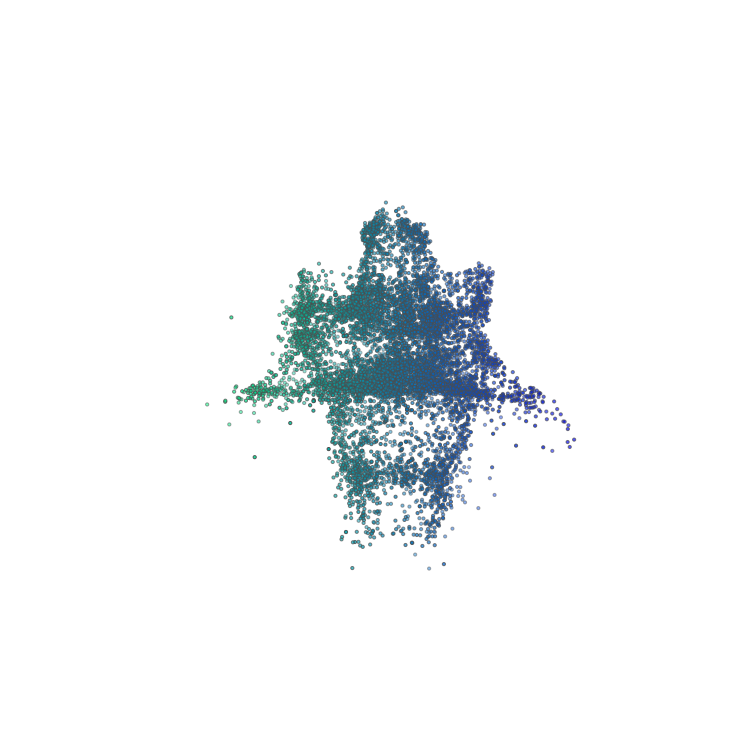}
        \caption{Input: cone, vase, lamp; \\ \qquad Target: person}
	\end{subfigure}
	\caption{ADD with dual and triple object input.}
	\label{fig:multiple}
\end{figure}

\paragraph{ADD with amalgamated inputs}In addition to using union during the transformation process, we can push this idea further by using amalgamated point clouds as input instead of a single object. We keep the experimental setting the same. Note that now we could have multiple of 10000 points as input. The results of ADD with the union objects in Figure~\ref{fig:union} are shown in Figure~\ref{fig:multiple}. Compared with Figure~\ref{fig:single}, ADD with multiple objects as input results in objects with more geometric details benefited from a more versatile input space. 

\subsection{Partitioned DeepDream (PDD)}

As the example shown in Figure~\ref{fig:deepdream_vis}, applying DeepDream to images produces more than one surreal patterns on each of the images~\cite{deepdream}. In contrast, the nature of ADD is to deform the whole object simultaneously into a certain shape. It is essential to extend such a basic algorithm in order to allow multiple, separate transformations on a single object. Therefore, we propose Partitioned DeepDream as shown in Algorithm~\ref{PDD} which can be implemented based on various point cloud segmentation algorithms. When compared with ADD, PDD provides a more controllable method for artists to explore with and effectively increases the diversity of creation.

\begin{algorithm}
	\caption{Partitioned DeepDream (PDD)}
  	\SetKwInOut{Input}{input}\SetKwInOut{Output}{output}
	\Input{trained $f_\theta$, number of segments $k$, input $X$ and targets $a$}
	\Output{generated object $\hat{X}$}
	$S=$ PCSegmentation$(X, k)$ \\
	\For{$x=S_1,\ldots,S_k$}{
	    Standardize x by $x = (x-\Bar{x})/\sigma_x$\\
	    $\hat{x} = ADD(f_\theta, x)$\\
	    Recover $\hat{x}$ by $\hat{x} = \hat{x}*\sigma_x+\Bar{x}$\\
	    $\hat{X} = \hat{X} \cup \hat{x}$
	}
	\label{PDD}
\end{algorithm}

\begin{figure}[!t]
    \begin{subfigure}{0.24\linewidth}
      \includegraphics[trim={100 155 100 55},clip,width=0.9\linewidth]{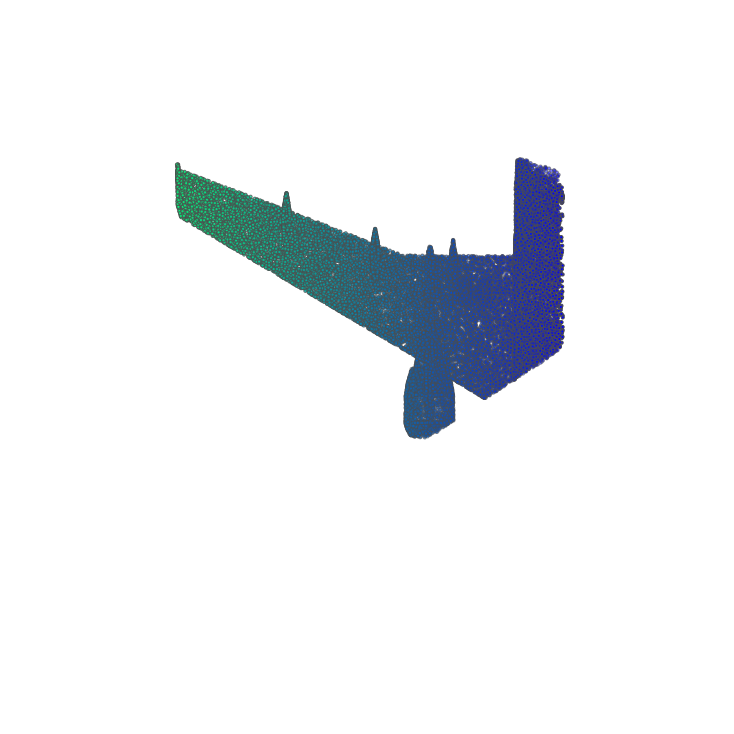}
    \end{subfigure}
    \begin{subfigure}{0.24\linewidth}
      \includegraphics[trim={100 55 100 155},clip,width=0.9\linewidth]{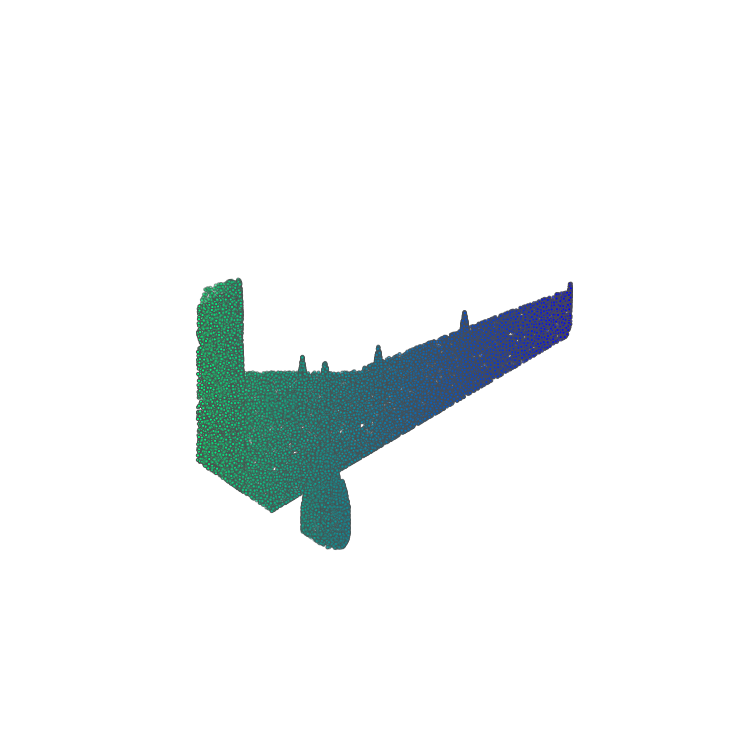}
    \end{subfigure}
    \begin{subfigure}{0.24\linewidth}
      \includegraphics[trim={100 100 100 100},clip,width=0.9\linewidth]{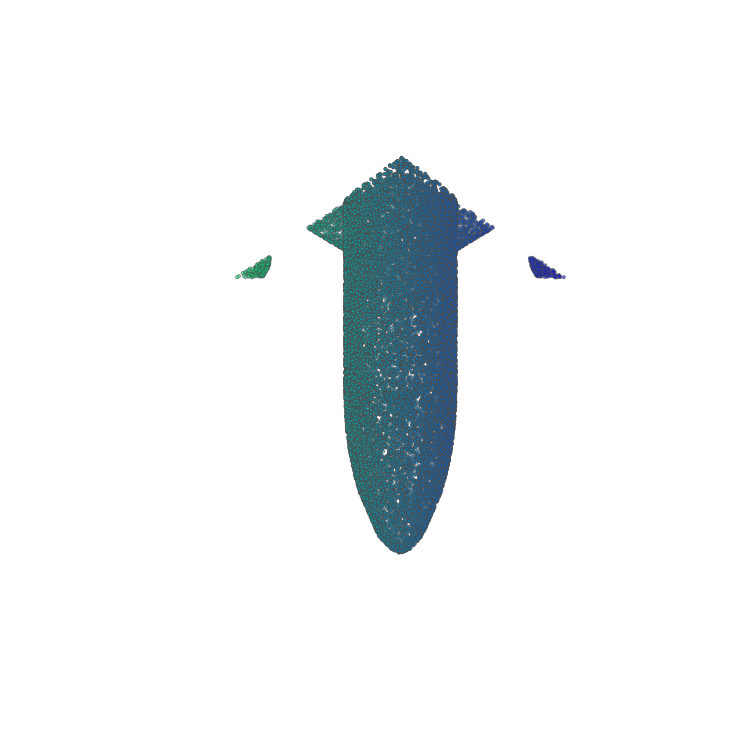}
    \end{subfigure}
    \begin{subfigure}{0.24\linewidth}
      \includegraphics[trim={100 100 100 100},clip,width=0.9\linewidth]{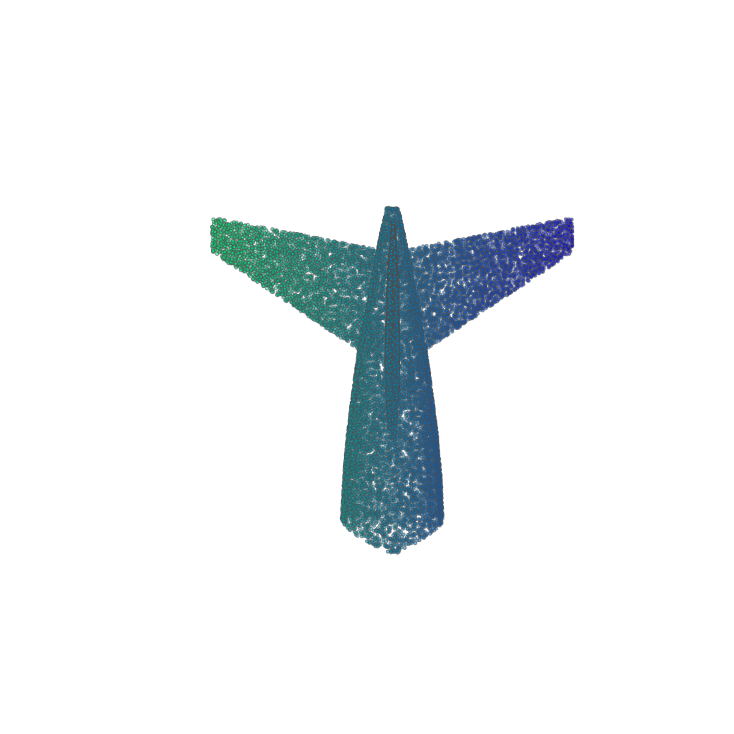}
    \end{subfigure}
    \caption{Segmentation results of an airplane with 4 clusters.}
    \label{fig:seg}
\end{figure}

\begin{figure}[!b]
    \begin{subfigure}{0.48\linewidth}
      \centering
      \includegraphics[trim={200 250 200 150},clip,width=0.9\linewidth]{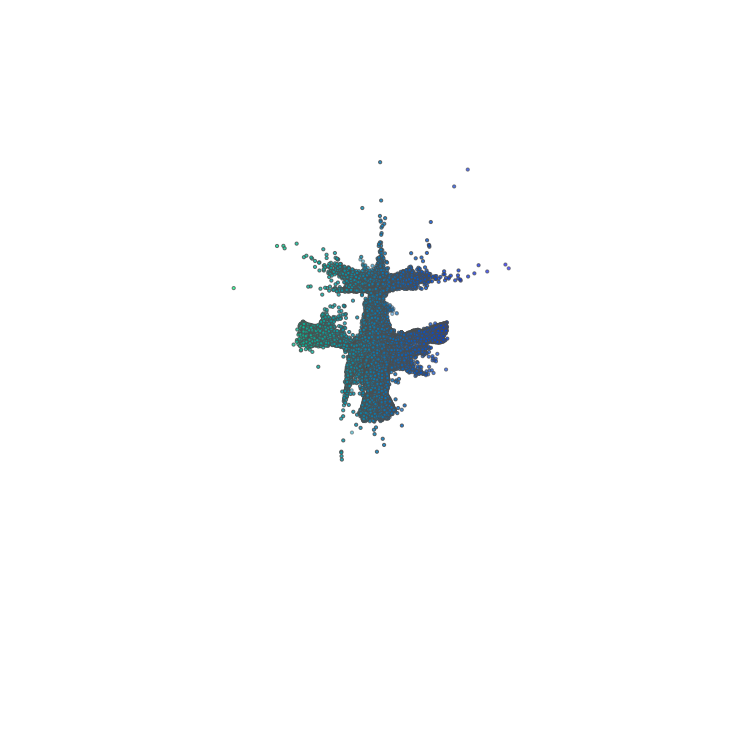}
      \caption{Input: airplane}
    \end{subfigure}
    \begin{subfigure}{0.48\linewidth}
      \centering
      \includegraphics[trim={200 200 200 200},clip,width=0.9\linewidth]{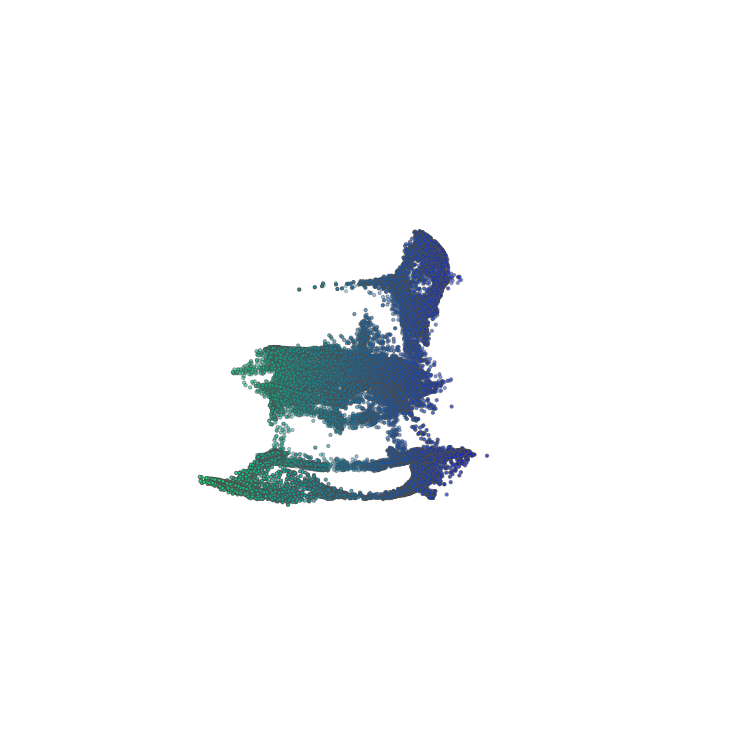}
      \caption{Input: chair}
    \end{subfigure}
    \caption{PDD targeting on random classes}
    \label{fig:kmeans}
\end{figure}

In contrast to the simplicity of mixing two point clouds, segmentation of point clouds is relatively hard and has been studied in depth~\cite{grilli2017review,nguyen20133d,trevor2013efficient}. Mainstream segmentation algorithms can be categorized into edge-based segmentation~\cite{sappa2001fast,wani2003parallel}, region growing segmentation~\cite{besl1988segmentation}, model-fitting segmentation~\cite{ballard1981generalizing,fischler1981random}, hybrid segmentation~\cite{vieira2005surface} and machine learning segmentation~\cite{lu2016pairwise}. In our work, manual segmentation is shown to obtain high-quality results but is extremely tedious and delegates too much responsibility to the artist. Instead, we explore machine learning methods which can automatically divide the object into several meaningful parts. For example, the segmentation results of an airplane with $k$-means are displayed in Figure \ref{fig:seg}. Note that this method requires the number of segments as input.

We extend ADD explained in Algorithm~\ref{algo:add} with the $k$-means point cloud segmentation algorithm in our experiments. First, the input object is automatically divided into several segments using $k$-means. Each segment must be standardized before it is fed into ADD, in order for it to be treated as one distinct object. With this preprocessing step ADD allows us to produce the desired features on different parts independently. In the end, we cancel the normalization and reunion all segments into one object.

\begin{figure}[!t]
    \begin{subfigure}{0.32\linewidth}
        \centering
        \includegraphics[trim={170 180 230 170},clip,width=0.9\linewidth]{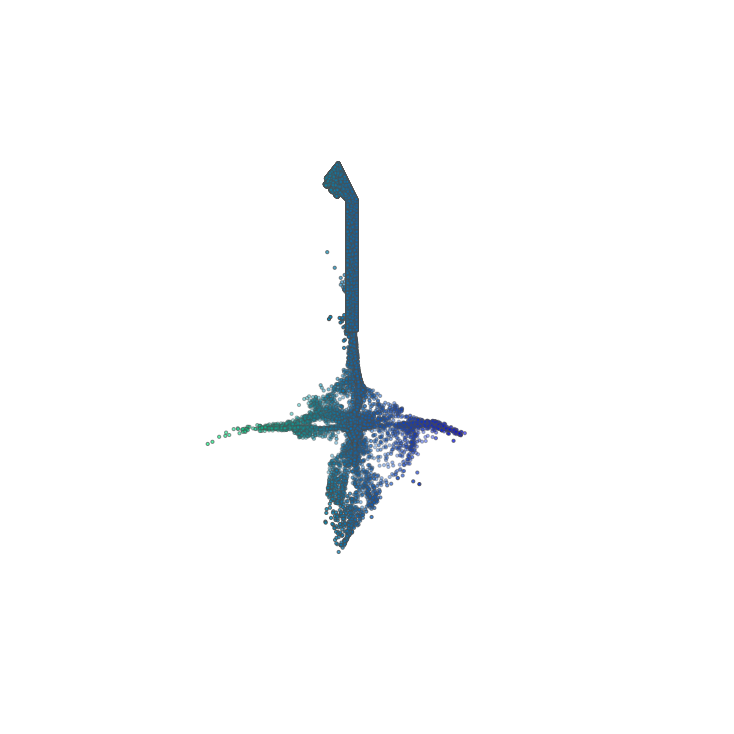}
        \caption{Target: airplane}
    \end{subfigure}
    \begin{subfigure}{0.32\linewidth}
        \centering
        \includegraphics[trim={200 180 200 170},clip,width=0.9\linewidth]{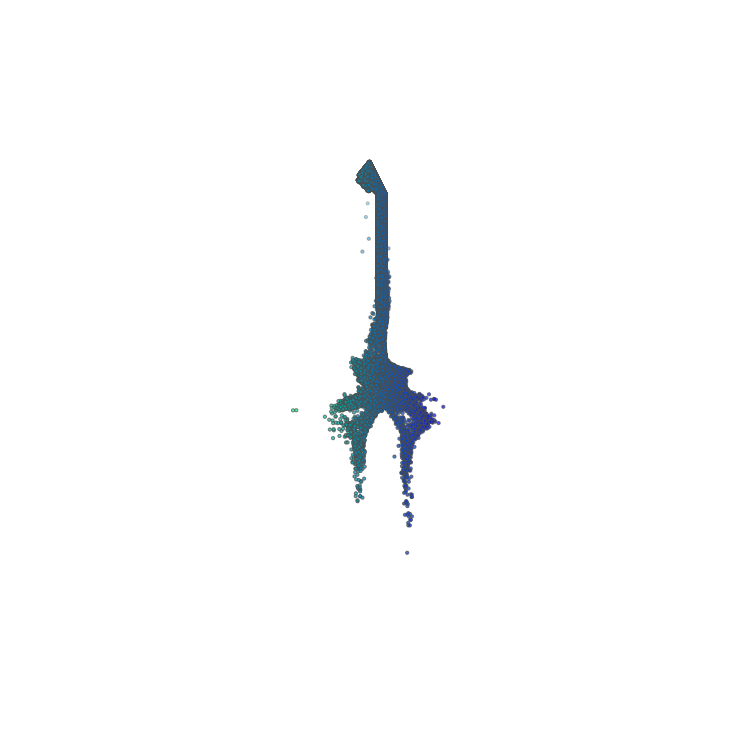}
        \caption{Target: person}
    \end{subfigure}
    \begin{subfigure}{0.32\linewidth}
        \centering
        \includegraphics[trim={150 180 250 170},clip,  width=0.9\linewidth]{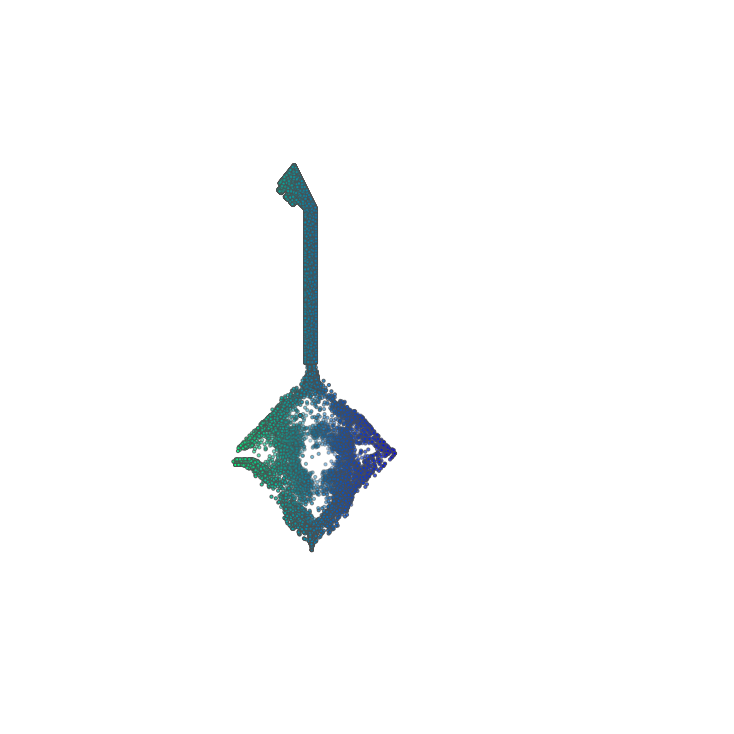}
        \caption{Target: bowl}
    \end{subfigure}
    \caption{PDD with guitar body only}
    \label{fig:PDD2}
\end{figure}

\begin{figure}[!b]
    \begin{subfigure}{0.48\linewidth}
        \centering
        \includegraphics[trim={150 200 275 200},clip,width=0.9\linewidth]{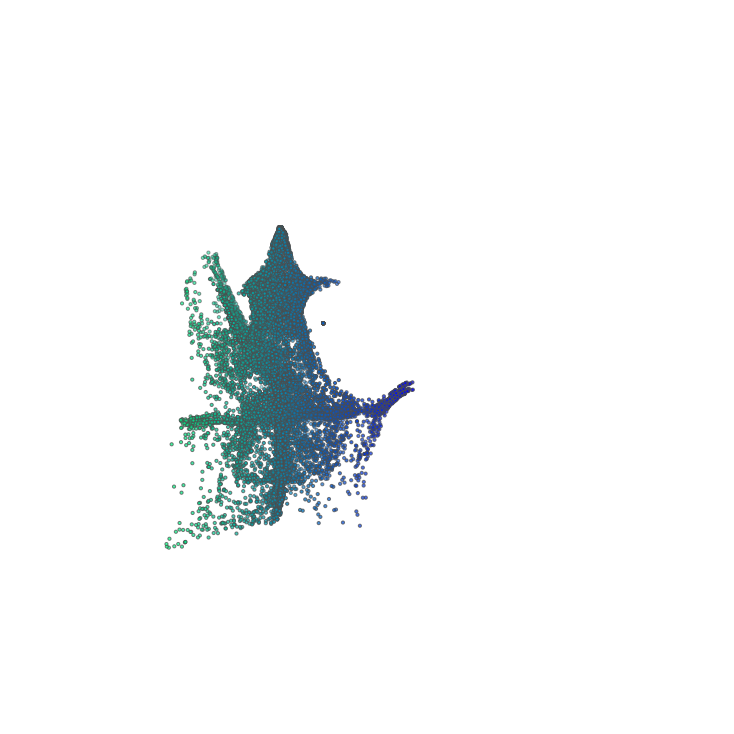}
        \caption{Input: bottle; Upper target: person; Lower: toilet}
    \end{subfigure}
    \begin{subfigure}{0.48\linewidth}
        \centering
        \includegraphics[trim={225 200 200 200},clip,width=0.9\linewidth]{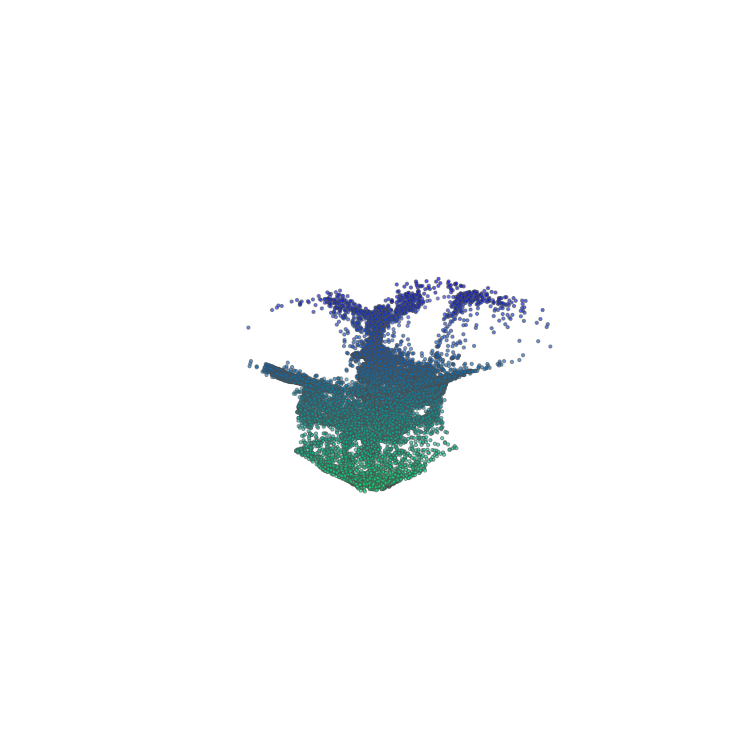}
        \caption{Input: ashcan; Upper target: cone; Lower: night\_stand}
    \end{subfigure}
    \caption{PDD on uniformly segmented objects}
    \label{fig:uniform}
\end{figure}

\paragraph{Experiments and results} The same data preprocessing methods and ADD model configurations are used here as mentioned in the aforementioned ADD experiments. As for the partition strategies, one can randomly select some targeted categories for different parts and PDD results are displayed in Figure~\ref{fig:kmeans}. This process is totally automatic with little human involvement. However, the random factors would lead to some undesired scattered points. Another more reasonable method is to create novel objects with more human participation by deforming the separate segments by design. As shown in Figure~\ref{fig:PDD2}, we create some point clouds from a guitar shown in Figure~\ref{fig:ShapeNet} by only hallucinating on their body parts, while keeping the neck part intact. As we can see, the created objects are only locally modified while the over layout remains that of a guitar.

\paragraph{PDD with manual segmentation strategies}
In addition to automatic segmentation algorithms, two manual segmentation strategies have been tested. First, for highly symmetric objects it is reasonable to uniformly divide it into several parts. As shown in Figure \ref{fig:uniform}, we manually partition a bottle and an ashcan from their middle, and run PDD with some selected targets on two segments separately. 

Another intuitive way to segment point clouds is to uniformly segment the whole spaces into many cubic blocks and treat the points in each block separately. The results based on this strategy are presented in Figure \ref{fig:block1} and \ref{fig:block2}. We partition the whole space containing the airplane into $3\times3\times2$ blocks alongside three axes, and generate features on each block with either a random class or cone class using PDD. In comparison to ADD, PDD can generate more delicate objects since it can manipulate different units of a single object. Besides, this also indicates a more expansive space for generating creative 3D objects. 

\begin{minipage}[!t]{\linewidth}
      \vspace{-1.4em}
      \begin{minipage}{0.48\linewidth}
          \begin{figure}[H]
              \begin{subfigure}[b]{.48\linewidth}
              \centering
              \includegraphics[trim={200 200 200 150},clip,width=\linewidth]{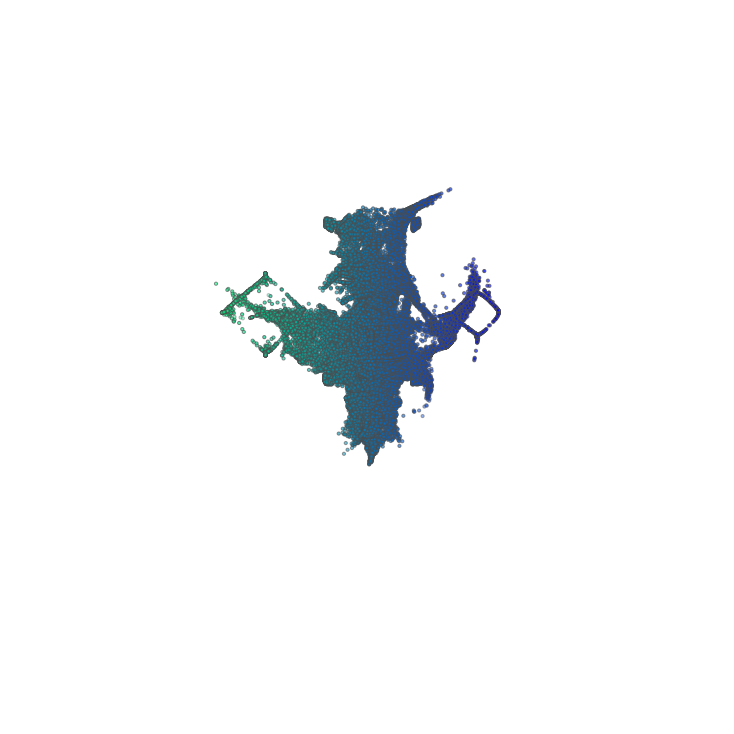}
              \end{subfigure}
              \begin{subfigure}[b]{.48\linewidth}
              \centering
              \includegraphics[trim={200 200 200 150},clip,width=\linewidth]{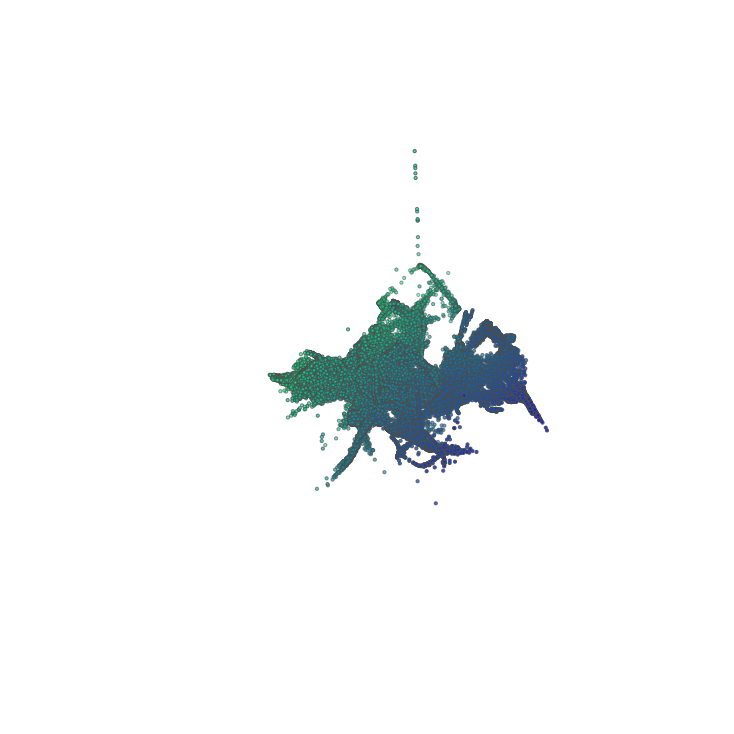}
              \end{subfigure}
              \caption{PDD with block segmented objects targeting random categories.}
              \label{fig:block1}
          \end{figure}
      \end{minipage}
      \hfill
      \begin{minipage}{0.48\linewidth}
          \begin{figure}[H]
              \centering
              \begin{subfigure}[b]{.48\linewidth}
              \centering
              \includegraphics[trim={180 200 220 150},clip,width=\linewidth]{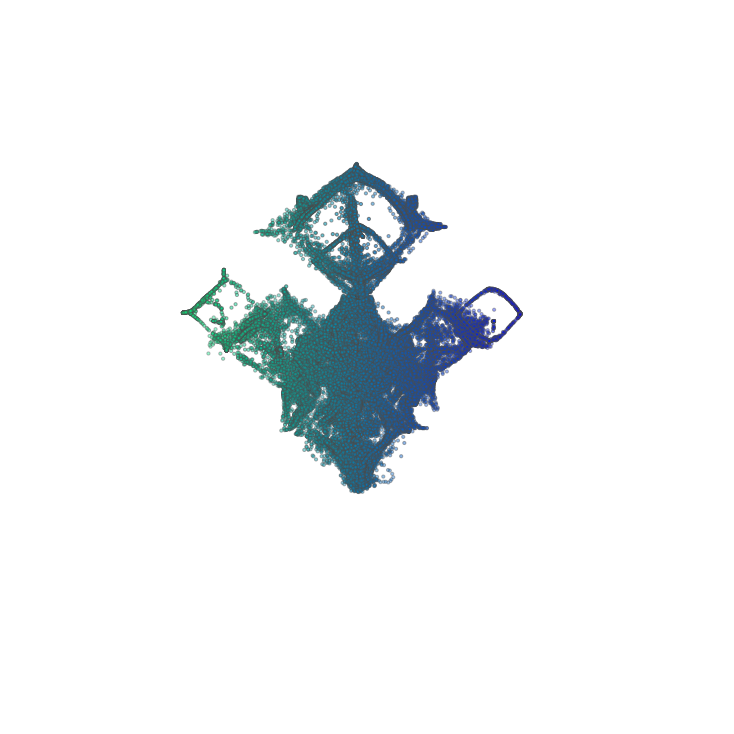}
              \end{subfigure}
              \begin{subfigure}[b]{.48\linewidth}
              \centering
              \includegraphics[trim={220 200 180 150},clip,width=\linewidth]{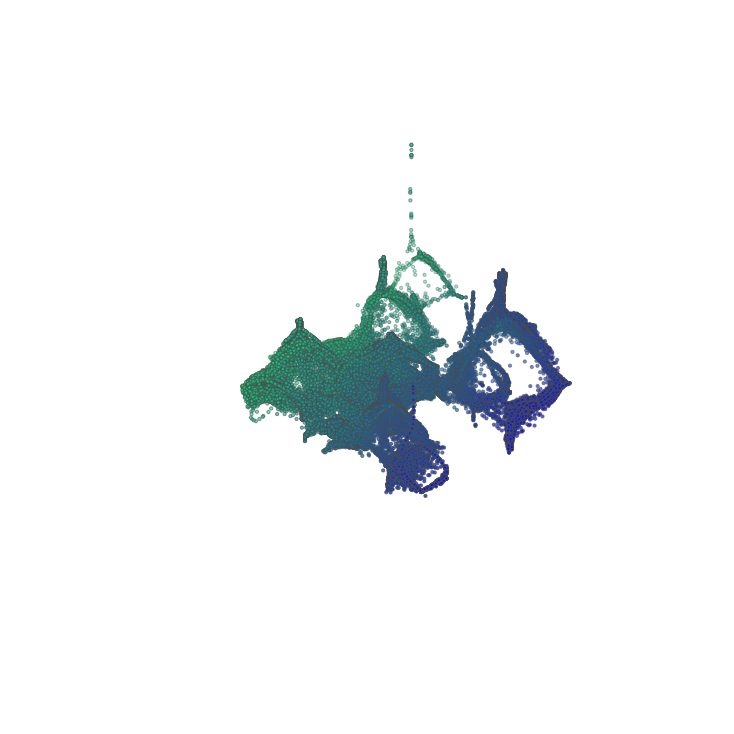}
              \end{subfigure}
              \caption{PDD with block segmented objects targeting cone.}
              \label{fig:block2}
          \end{figure}
      \end{minipage}
      \hfill
\end{minipage}

\section{Realization of Sculptural Object}

\paragraph{From points clouds to mesh}

Given a novel point cloud, our task becomes generating a mesh -- a description of an object using vertices, edges, and faces -- in order to eliminate the spatial ambiguity of a point cloud representation. While mesh generation, also known as surface reconstruction, is readily provided in the free software package MeshLab~\citep{meshlab}, each possible algorithm has its own use case, so this task involves experimenting with each available algorithm to find what generates the best reconstruction for a given point cloud. 

Our experiments focused on the following reconstruction algorithms: Delaunay Triangulation~\cite{cazals2006delaunay}, Screened Poisson~\citep{kazhdan2013screened}, Alpha Complex~\citep{guo1997surface}, and Ball-Pivoting~\citep{bernardini1999ball}. While each of these algorithms has its uses, ADD results in point clouds with a greater variance in sparsity than typical point clouds generated from real-world objects.

\begin{figure}[!t]
\centering
\captionsetup[subfigure]{justification=centering}
\begin{subfigure}{.48\linewidth}
  \centering
  \includegraphics[width=.9\linewidth]{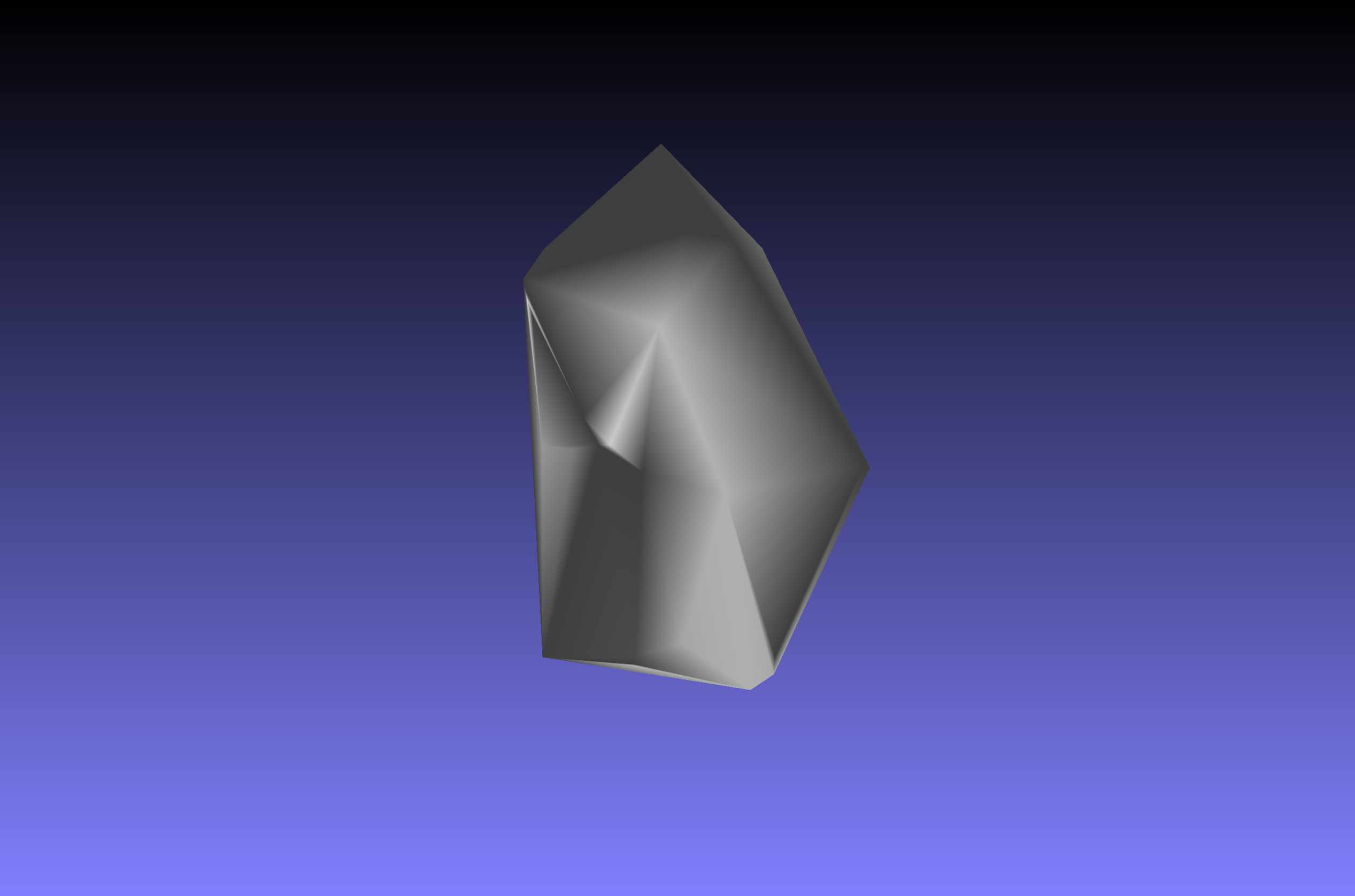}
  \caption{Delaunay \\ Triangulation}
  \label{fig:delaunay}
\end{subfigure}%
\begin{subfigure}{.48\linewidth}
  \centering
  \includegraphics[width=.9\linewidth]{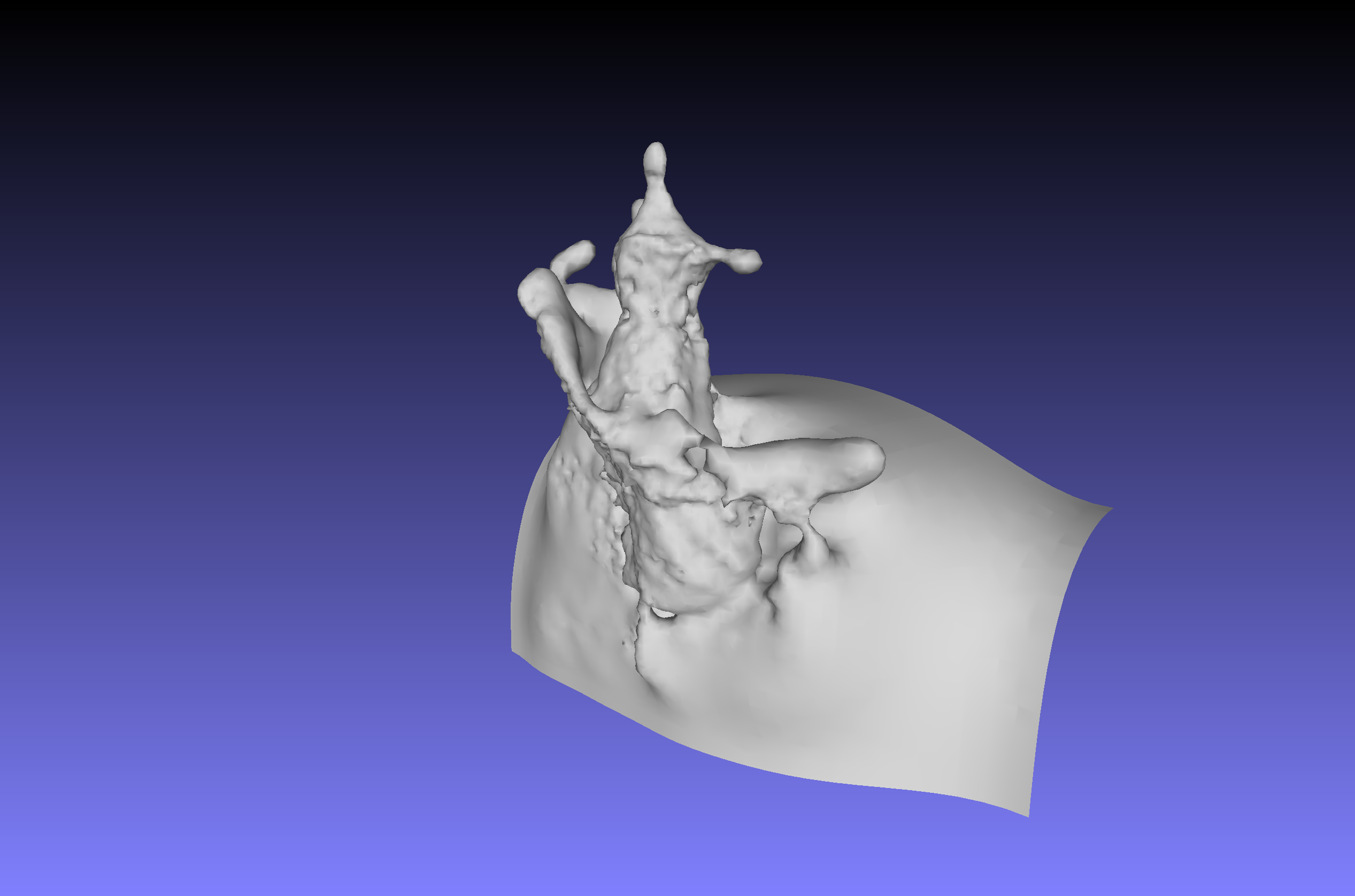}
  \caption{Screened \\ Poisson}
  \label{fig:poisson}
\end{subfigure}
\begin{subfigure}{.48\linewidth}
  \centering
  \includegraphics[width=.9\linewidth]{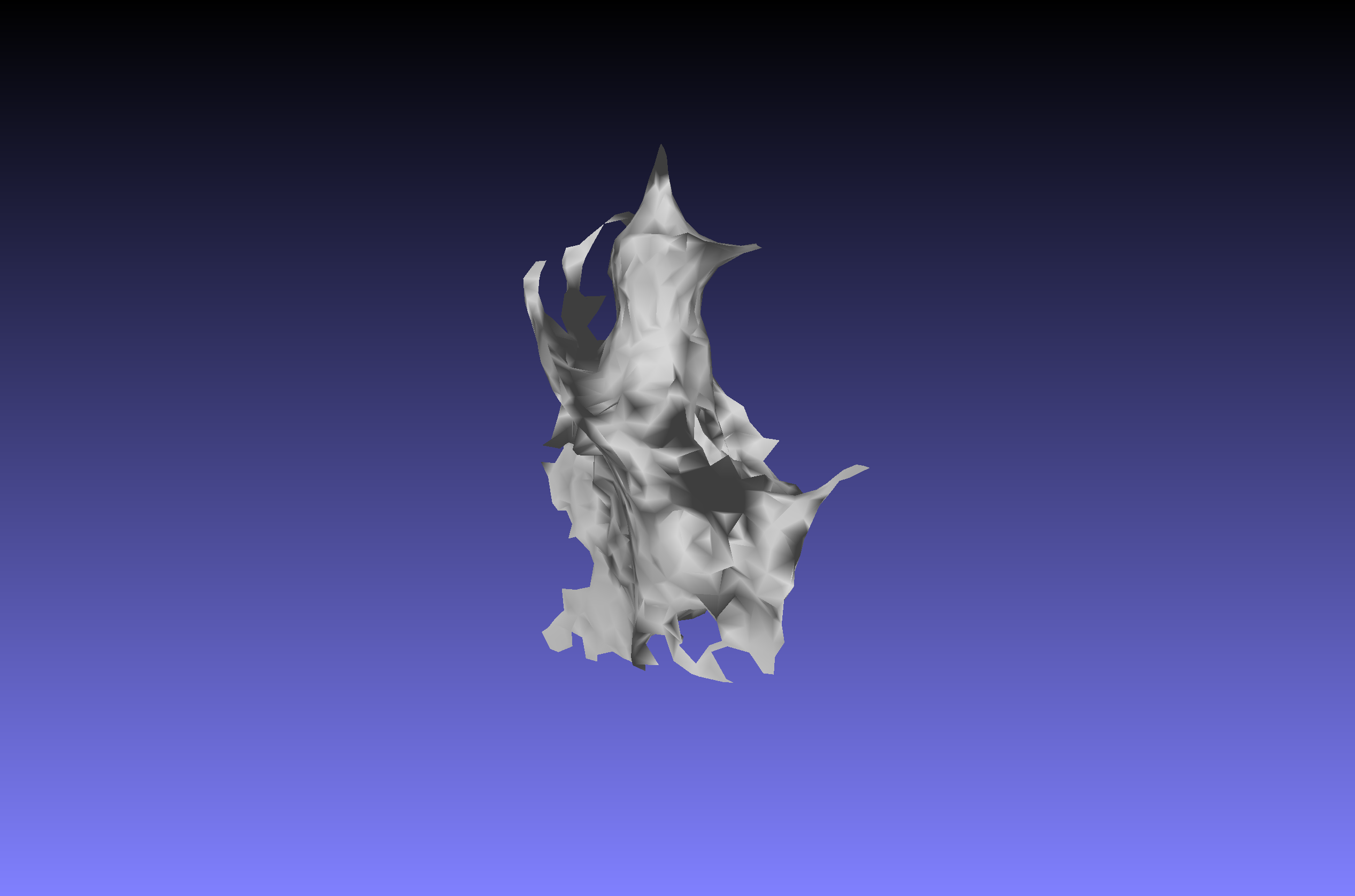}
  \caption{Alpha \\ Complex}
  \label{fig:alpha}
\end{subfigure}
\begin{subfigure}{.48\linewidth}
  \centering
  \includegraphics[width=.9\linewidth]{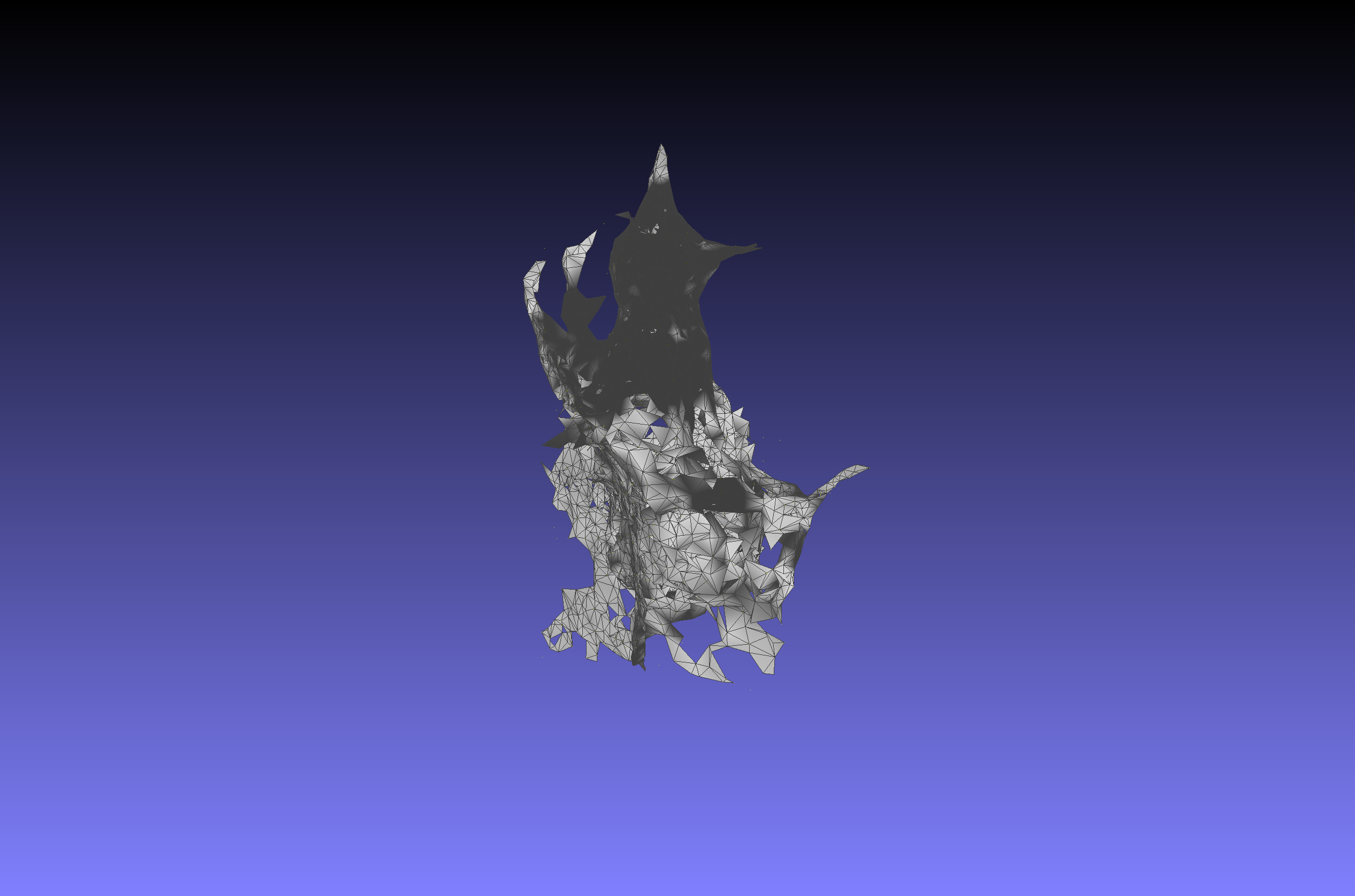}
  \caption{Ball \\ Pivoting}
  \label{fig:bp}
\end{subfigure}
\caption{Reconstruction techniques}
\label{fig:reconstruction}
\end{figure}

For all experiments, we limited the number of points to 10,000. If a generated point cloud had more than this number, we reduced using Poisson Disk Sub-sampling~\citep{poissondisk}. This sub-sampling method is more intelligent than a simple uniform resampling and produces points that appear to balance sparsity with finer definition. 

An illustration of our findings can be found in Figure \ref{fig:reconstruction}. Typically, Delaunay Triangulation produces a mesh that eliminates all detail of the point cloud. This approach will therefore only be appropriate when it is applied to an exceptionally sparse point cloud, as it would contain minimal detail regardless of the reconstruction technique. Screened Poisson extrapolates from the original point cloud, producing interesting shapes that are unfortunately misleading of the underlying structure. Alpha Complex and Ball-Pivoting are the two most consistent reconstruction methods and they both produce similar meshes. One important difference is that Alpha Complex produces non-manifold edges, all but making faithful 3D printing impossible.

As a result of these trials, we settled on Ball-Pivoting after Poisson Disk Sub-sampling to produce our final mesh.

\begin{figure}[!b]
    \centering
    \includegraphics[width=0.6\linewidth]{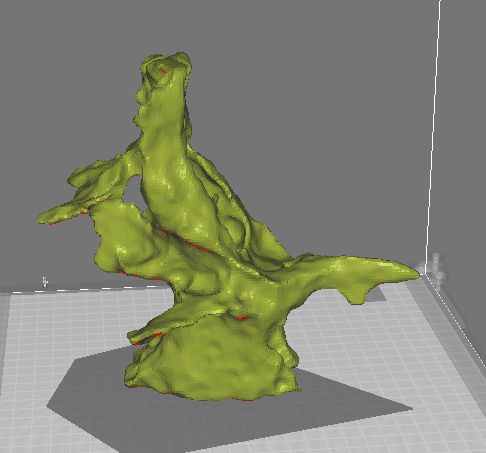}
    \caption{Screen capture from Cura-Lulzbot software for 3d printing}
    \label{fig:printable_3d}
\end{figure}

\begin{figure}[!h]
    \begin{subfigure}{.48\linewidth}
        \centering
        \includegraphics[width=0.9\linewidth]{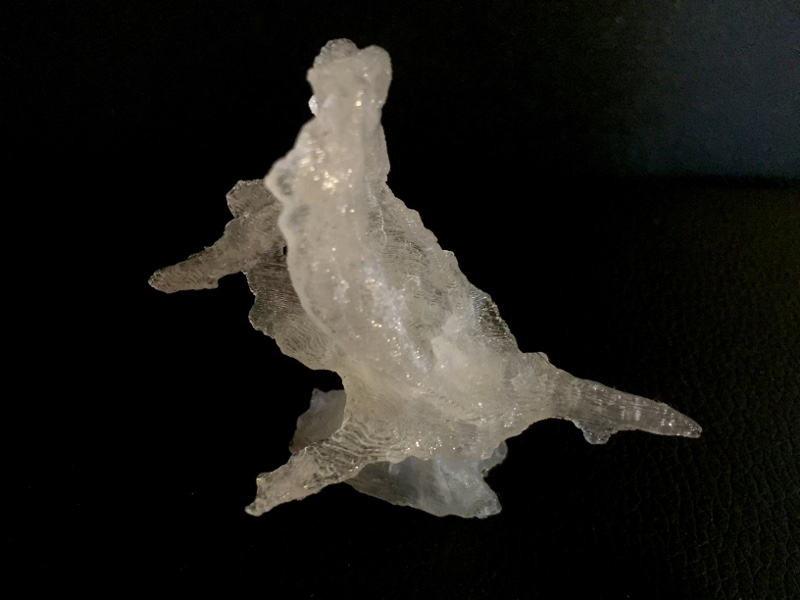}
    \end{subfigure}
	\begin{subfigure}{.48\linewidth}
	    \centering
        \includegraphics[width=0.9\linewidth]{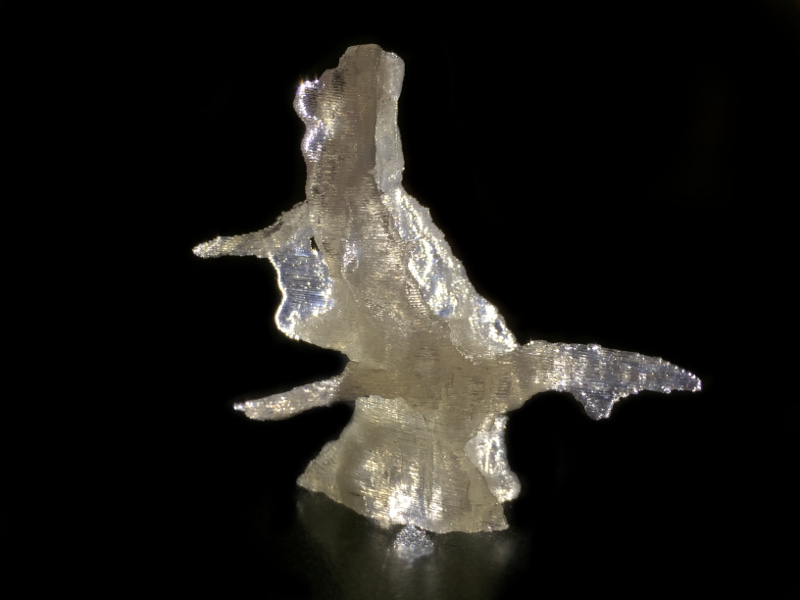}
	\end{subfigure}
	\caption{Created sculpture from ADD.}
	\label{fig:print}
\end{figure}

\paragraph{From mesh to realizable 3D sculptures}
Our final goal is to create 3D sculptures in the real world. 
We use standard software Meshmixer ~\citep{schmidt2010meshmixer} to solidify the surfaced reconstructed point cloud
created by ADD. The solidified version of Figure \ref{fig:bp} can be seen in Figure \ref{fig:print_pdd}. We then use Lulzbot TAZ 3D printer with dual-extruder, transparent PLA, and dissolvable supporting material to create the sculpture of the reconstructed mesh. The printed sculptures of the point cloud in Figure~\ref{fig:multiple} (RHS) are shown in Figure~\ref{fig:print}.

\begin{figure}[!h]
    \begin{subfigure}{.49\linewidth}
        \centering
        \includegraphics[width=\linewidth]{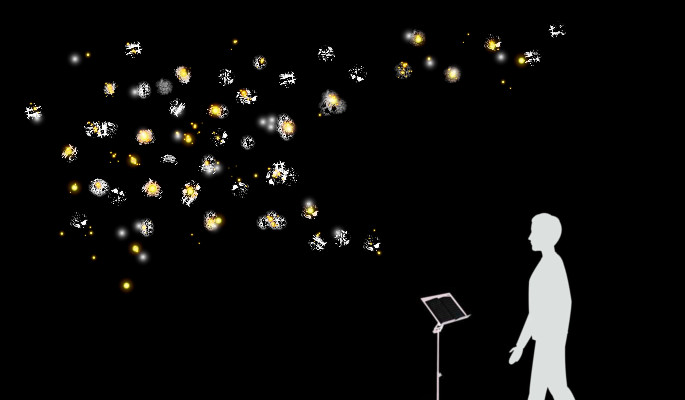}
    \end{subfigure}
    \begin{subfigure}{.47\linewidth}
        \centering
        \includegraphics[width=\linewidth]{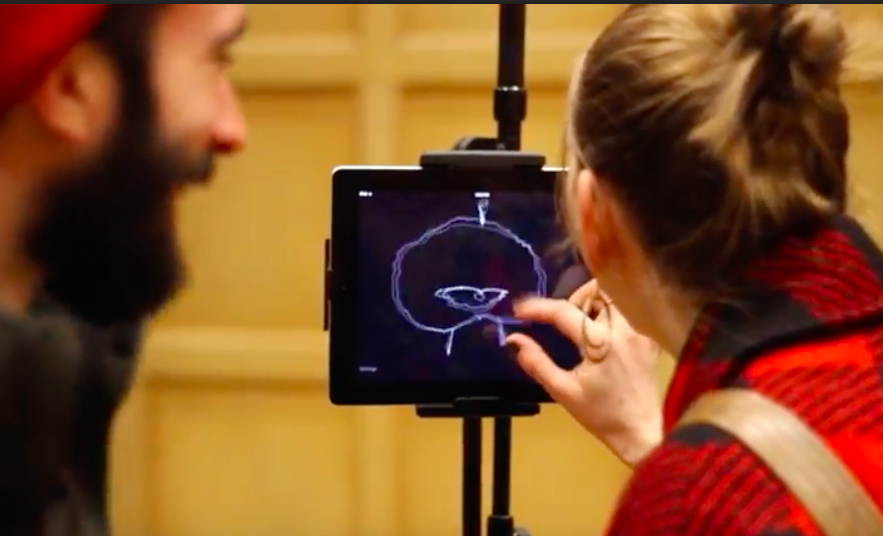} 
    \end{subfigure}
    \caption{Sketch of Aural Fauna: Illuminato art installation and a Photo from the touch-sound interaction test with tablet computer interface}
    \label{fig:auralfauna}
\end{figure}
\vspace{-0.5cm}

\section{Conclusion}

This paper describes our development of two ML algorithms for point cloud generation, ADD and PDD, which possess the ability to generate unique sculptural artwork, distinct from the dataset used to train the underlying model. These approaches allow more flexibility in the deformation in the underlying object as well as the ability to control the density of points required for meshing and printing 3D sculptures. 

Through our development of ADD and PDD for sculptural object generation, we show AI's potential for creative endeavours in art, especially the traditional field of sculpture and installation art. One challenge in the development of creative AI is the contradiction between the need of an objective metric for evaluating the quality of generated objects in the ML literature and the desire for creative and expressive forms in the art community. Reconciling these goals will require a mathematical definition of creativity that can help guide the AI's generative process for each specific case. We will undertake these deep computational and philosophical questions in future work. 

In the meantime, the realized 3D objects are an integral part of our ongoing project that is an interactive art installation. This project, Aural Fauna: Illuminato, presents an unknown form of organism, aural fauna, that reacts to the visitor’s sound and touch. Their sound is generated using machine learning. The creatures’ bodies are the generated sculptural objects by ADD and PDD. The sculptural objects could be installed on the wall or be hung from the ceiling like a swarm and the participant may interact with them using the touch-screen interface and their voice input as seen in Figure~\ref{fig:auralfauna}. This project aims to encapsulate the joint efforts of human artists and Creative AI as well as the viewer as partners in the endeavor. 


\bibliographystyle{isea}
\bibliography{mybib}

\begin{thebibliography}{}

\bibitem[\protect\citeauthoryear{Achlioptas \bgroup et al.\egroup
  }{2017}]{achlioptas2017learning}
Achlioptas, P.; Diamanti, O.; Mitliagkas, I.; and Guibas, L.
\newblock 2017.
\newblock Learning representations and generative models for 3d point clouds.
\newblock {\em arXiv preprint arXiv:1707.02392}.

\bibitem[\protect\citeauthoryear{Ballard}{1981}]{ballard1981generalizing}
Ballard, D.~H.
\newblock 1981.
\newblock Generalizing the hough transform to detect arbitrary shapes.
\newblock {\em Pattern recognition} 13(2):111--122.

\bibitem[\protect\citeauthoryear{Bernardini \bgroup et al.\egroup
  }{1999}]{bernardini1999ball}
Bernardini, F.; Mittleman, J.; Rushmeier, H.; Silva, C.; and Taubin, G.
\newblock 1999.
\newblock The ball-pivoting algorithm for surface reconstruction.
\newblock {\em IEEE transactions on visualization and computer graphics}
  5(4):349--359.

\bibitem[\protect\citeauthoryear{Besl and Jain}{1988}]{besl1988segmentation}
Besl, P.~J., and Jain, R.~C.
\newblock 1988.
\newblock Segmentation through variable-order surface fitting.
\newblock {\em IEEE Transactions on Pattern Analysis and Machine Intelligence}
  10(2):167--192.

\bibitem[\protect\citeauthoryear{Bidgoli and Veloso}{2018}]{DeepCloud}
Bidgoli, A., and Veloso, P.
\newblock 2018.
\newblock Deepcloud: The application of a data-driven, generative model in
  design.

\bibitem[\protect\citeauthoryear{{Brown} \bgroup et al.\egroup
  }{2018}]{unrestricted_Advex_2018}
{Brown}, T.~B.; {Carlini}, N.; {Zhang}, C.; {Olsson}, C.; {Christiano}, P.; and
  {Goodfellow}, I.
\newblock 2018.
\newblock Unrestricted adversarial examples.
\newblock {\em arXiv preprint arXiv:1809.08352}.

\bibitem[\protect\citeauthoryear{Carter and Nielsen}{2017}]{carter2017using}
Carter, S., and Nielsen, M.
\newblock 2017.
\newblock Using artificial intelligence to augment human intelligence.
\newblock {\em Distill}.

\bibitem[\protect\citeauthoryear{Cazals and Giesen}{2006}]{cazals2006delaunay}
Cazals, F., and Giesen, J.
\newblock 2006.
\newblock Delaunay triangulation based surface reconstruction.
\newblock In {\em Effective computational geometry for curves and surfaces}.
  Springer.
\newblock  231--276.

\bibitem[\protect\citeauthoryear{Chang \bgroup et al.\egroup
  }{2015}]{chang2015shapenet}
Chang, A.~X.; Funkhouser, T.; Guibas, L.; Hanrahan, P.; Huang, Q.; Li, Z.;
  Savarese, S.; Savva, M.; Song, S.; Su, H.; et~al.
\newblock 2015.
\newblock Shapenet: An information-rich 3d model repository.
\newblock {\em arXiv preprint arXiv:1512.03012}.

\bibitem[\protect\citeauthoryear{Cignoni \bgroup et al.\egroup
  }{2008}]{meshlab}
Cignoni, P.; Callieri, M.; Corsini, M.; Dellepiane, M.; Ganovelli, F.; and
  Ranzuglia, G.
\newblock 2008.
\newblock {MeshLab: an Open-Source Mesh Processing Tool}.
\newblock In Scarano, V.; Chiara, R.~D.; and Erra, U., eds., {\em Eurographics
  Italian Chapter Conference}.
\newblock The Eurographics Association.

\bibitem[\protect\citeauthoryear{con}{}]{contentaware}
Content aware studies.
\newblock http://egorkraft.art/\#cas.

\bibitem[\protect\citeauthoryear{Corsini, Cignoni, and
  Scopigno}{2012}]{poissondisk}
Corsini, M.; Cignoni, P.; and Scopigno, R.
\newblock 2012.
\newblock Efficient and flexible sampling with blue noise properties of
  triangular meshes.
\newblock {\em IEEE Transaction on Visualization and Computer Graphics}
  18(6):914--924.
\newblock http://doi.ieeecomputersociety.org/10.1109/TVCG.2012.34.

\bibitem[\protect\citeauthoryear{Elgammal \bgroup et al.\egroup
  }{2017}]{elgammal2017can}
Elgammal, A.; Liu, B.; Elhoseiny, M.; and Mazzone, M.
\newblock 2017.
\newblock Can: Creative adversarial networks, generating" art" by learning
  about styles and deviating from style norms.
\newblock {\em arXiv preprint arXiv:1706.07068}.

\bibitem[\protect\citeauthoryear{Fischler and
  Bolles}{1981}]{fischler1981random}
Fischler, M.~A., and Bolles, R.~C.
\newblock 1981.
\newblock Random sample consensus: a paradigm for model fitting with
  applications to image analysis and automated cartography.
\newblock {\em Communications of the ACM} 24(6):381--395.

\bibitem[\protect\citeauthoryear{Goodfellow \bgroup et al.\egroup
  }{2014}]{goodfellow2014generative}
Goodfellow, I.; Pouget-Abadie, J.; Mirza, M.; Xu, B.; Warde-Farley, D.; Ozair,
  S.; Courville, A.; and Bengio, Y.
\newblock 2014.
\newblock Generative adversarial nets.
\newblock In {\em NIPS}.

\bibitem[\protect\citeauthoryear{Grilli, Menna, and
  Remondino}{2017}]{grilli2017review}
Grilli, E.; Menna, F.; and Remondino, F.
\newblock 2017.
\newblock A review of point clouds segmentation and classification algorithms.
\newblock {\em The International Archives of Photogrammetry, Remote Sensing and
  Spatial Information Sciences} 42:339.

\bibitem[\protect\citeauthoryear{Groueix \bgroup et al.\egroup
  }{2018}]{groueix2018atlasnet}
Groueix, T.; Fisher, M.; Kim, V.~G.; Russell, B.~C.; and Aubry, M.
\newblock 2018.
\newblock Atlasnet: A papier-mache approach to learning 3d surface generation.
\newblock {\em arXiv preprint arXiv:1802.05384}.

\bibitem[\protect\citeauthoryear{Guo, Menon, and
  Willette}{1997}]{guo1997surface}
Guo, B.; Menon, J.; and Willette, B.
\newblock 1997.
\newblock Surface reconstruction using alpha shapes.
\newblock In {\em Computer Graphics Forum}, volume~16,  177--190.
\newblock Wiley Online Library.

\bibitem[\protect\citeauthoryear{Hoppe \bgroup et al.\egroup
  }{1992}]{hoppe1992surface}
Hoppe, H.; DeRose, T.; Duchamp, T.; McDonald, J.; and Stuetzle, W.
\newblock 1992.
\newblock {\em Surface reconstruction from unorganized points}.

\bibitem[\protect\citeauthoryear{Kato, Ushiku, and
  Harada}{2018}]{kato2018neural}
Kato, H.; Ushiku, Y.; and Harada, T.
\newblock 2018.
\newblock Neural 3d mesh renderer.
\newblock In {\em CVPR}.

\bibitem[\protect\citeauthoryear{Kazhdan and Hoppe}{2013}]{kazhdan2013screened}
Kazhdan, M., and Hoppe, H.
\newblock 2013.
\newblock Screened poisson surface reconstruction.
\newblock {\em ACM Transactions on Graphics (TOG)} 32(3):29.

\bibitem[\protect\citeauthoryear{Lehman, Risi, and
  Clune}{2016}]{lehman2016creative}
Lehman, J.; Risi, S.; and Clune, J.
\newblock 2016.
\newblock Creative generation of 3d objects with deep learning and innovation
  engines.
\newblock In {\em Proceedings of the 7th International Conference on
  Computational Creativity}.

\bibitem[\protect\citeauthoryear{Li \bgroup et al.\egroup }{2018}]{li2018point}
Li, C.-L.; Zaheer, M.; Zhang, Y.; Poczos, B.; and Salakhutdinov, R.
\newblock 2018.
\newblock Point cloud gan.
\newblock {\em arXiv preprint arXiv:1810.05795}.

\bibitem[\protect\citeauthoryear{Liu \bgroup et al.\egroup
  }{2018}]{liu2018adversarial}
Liu, H.-T.~D.; Tao, M.; Li, C.-L.; Nowrouzezahrai, D.; and Jacobson, A.
\newblock 2018.
\newblock Adversarial geometry and lighting using a differentiable renderer.
\newblock {\em arXiv preprint arXiv:1808.02651}.

\bibitem[\protect\citeauthoryear{Lu \bgroup et al.\egroup
  }{2016}]{lu2016pairwise}
Lu, X.; Yao, J.; Tu, J.; Li, K.; Li, L.; and Liu, Y.
\newblock 2016.
\newblock Pairwise linkage for point cloud segmentation.
\newblock {\em ISPRS Annals of Photogrammetry, Remote Sensing \& Spatial
  Information Sciences} 3(3).

\bibitem[\protect\citeauthoryear{Nguyen and Le}{2013}]{nguyen20133d}
Nguyen, A., and Le, B.
\newblock 2013.
\newblock 3d point cloud segmentation: A survey.
\newblock In {\em RAM},  225--230.

\bibitem[\protect\citeauthoryear{Nima, Luca, and Mauro}{2018}]{nima2018vox2net}
Nima, D.; Luca, S.; and Mauro, M.
\newblock 2018.
\newblock Vox2net: From 3d shapes to network sculptures.
\newblock In {\em NIPS}.

\bibitem[\protect\citeauthoryear{Posted~by Alexander~Mordvintsev and
  Tyka}{}]{deepdream}
Posted~by Alexander~Mordvintsev, C.~O., and Tyka, M.
\newblock Inceptionism: Going deeper into neural networks.
\newblock
  https://ai.googleblog.com/2015/06/inceptionism-going-deeper-into-neural.html.

\bibitem[\protect\citeauthoryear{Qi \bgroup et al.\egroup
  }{2017}]{qi2017pointnet}
Qi, C.~R.; Su, H.; Mo, K.; and Guibas, L.~J.
\newblock 2017.
\newblock Pointnet: Deep learning on point sets for 3d classification and
  segmentation.
\newblock {\em CVPR}.

\bibitem[\protect\citeauthoryear{Robert}{2014}]{robert2014machine}
Robert, C.
\newblock 2014.
\newblock Machine learning, a probabilistic perspective.

\bibitem[\protect\citeauthoryear{Sappa and Devy}{2001}]{sappa2001fast}
Sappa, A.~D., and Devy, M.
\newblock 2001.
\newblock Fast range image segmentation by an edge detection strategy.
\newblock In {\em 3-D Digital Imaging and Modeling, 2001. Proceedings. Third
  International Conference on},  292--299.
\newblock IEEE.

\bibitem[\protect\citeauthoryear{Schmidt and
  Singh}{2010}]{schmidt2010meshmixer}
Schmidt, R., and Singh, K.
\newblock 2010.
\newblock Meshmixer: an interface for rapid mesh composition.
\newblock In {\em ACM SIGGRAPH 2010 Talks}, ~6.
\newblock ACM.

\bibitem[\protect\citeauthoryear{Szegedy \bgroup et al.\egroup
  }{2013}]{szegedy2013intriguing}
Szegedy, C.; Zaremba, W.; Sutskever, I.; Bruna, J.; Erhan, D.; Goodfellow, I.;
  and Fergus, R.
\newblock 2013.
\newblock Intriguing properties of neural networks.
\newblock {\em arXiv preprint arXiv:1312.6199}.

\bibitem[\protect\citeauthoryear{Trevor \bgroup et al.\egroup
  }{2013}]{trevor2013efficient}
Trevor, A.~J.; Gedikli, S.; Rusu, R.~B.; and Christensen, H.~I.
\newblock 2013.
\newblock Efficient organized point cloud segmentation with connected
  components.
\newblock {\em Semantic Perception Mapping and Exploration (SPME)}.

\bibitem[\protect\citeauthoryear{Van Den~Oord \bgroup et al.\egroup
  }{2016}]{van2016wavenet}
Van Den~Oord, A.; Dieleman, S.; Zen, H.; Simonyan, K.; Vinyals, O.; Graves, A.;
  Kalchbrenner, N.; Senior, A.~W.; and Kavukcuoglu, K.
\newblock 2016.
\newblock Wavenet: A generative model for raw audio.
\newblock In {\em SSW}.

\bibitem[\protect\citeauthoryear{Vieira and Shimada}{2005}]{vieira2005surface}
Vieira, M., and Shimada, K.
\newblock 2005.
\newblock Surface mesh segmentation and smooth surface extraction through
  region growing.
\newblock {\em Computer aided geometric design} 22(8):771--792.

\bibitem[\protect\citeauthoryear{Wani and Arabnia}{2003}]{wani2003parallel}
Wani, M.~A., and Arabnia, H.~R.
\newblock 2003.
\newblock Parallel edge-region-based segmentation algorithm targeted at
  reconfigurable multiring network.
\newblock {\em The Journal of Supercomputing} 25(1):43--62.

\bibitem[\protect\citeauthoryear{Wu \bgroup et al.\egroup }{2015}]{wu20153d}
Wu, Z.; Song, S.; Khosla, A.; Yu, F.; Zhang, L.; Tang, X.; and Xiao, J.
\newblock 2015.
\newblock 3d shapenets: A deep representation for volumetric shapes.
\newblock In {\em CVPR}.

\bibitem[\protect\citeauthoryear{Wu \bgroup et al.\egroup
  }{2016}]{wu2016learning}
Wu, J.; Zhang, C.; Xue, T.; Freeman, B.; and Tenenbaum, J.
\newblock 2016.
\newblock Learning a probabilistic latent space of object shapes via 3d
  generative-adversarial modeling.
\newblock In {\em NIPS}.

\bibitem[\protect\citeauthoryear{Yang \bgroup et al.\egroup
  }{2018}]{yang2018foldingnet}
Yang, Y.; Feng, C.; Shen, Y.; and Tian, D.
\newblock 2018.
\newblock Foldingnet: Point cloud auto-encoder via deep grid deformation.
\newblock In {\em CVPR}, volume~3.

\bibitem[\protect\citeauthoryear{Zaheer \bgroup et al.\egroup
  }{2017}]{zaheer2017deep}
Zaheer, M.; Kottur, S.; Ravanbakhsh, S.; Poczos, B.; Salakhutdinov, R.~R.; and
  Smola, A.~J.
\newblock 2017.
\newblock Deep sets.
\newblock In {\em NIPS}.

\end{thebibliography}

\section{Authors Biographies}

Songwei Ge is a Masters student in the Computational Biology Department at Carnegie Mellon University.

Austin Dill is a Masters student in the Machine Learning Department at Carnegie Mellon University.

Chun-Liang Li is a PhD candidate in the Machine Learning Department of Carnegie Mellon University. He received IBM Ph.D. Fellowship in 2018 and was the Best Student Paper runner-up at the International Joint Conference on Artificial Intelligence (IJCAI) in 2017. His research interest is on deep generative models from theories to practical applications.

Dr. Eunsu Kang is a Korean media artist who creates interactive audiovisual installations and AI artworks. Her current research is focused on creative AI and artistic expressions generated by Machine Learning algorithms. Creating interdisciplinary projects, her signature has been seamless integration of art disciplines and innovative techniques. Her work has been invited to numerous places around the world including Korea, Japan, China, Switzerland, Sweden, France, Germany, and the US. All ten of her solo shows, consisting of individual or collaborative projects, were invited or awarded. She has won the Korean National Grant for Arts three times. Her researches have been presented at prestigious conferences including ACM, ICMC, ISEA, and NeurIPS. Kang earned her Ph.D. in Digital Arts and Experimental Media from DXARTS at the University of Washington. She received an MA in Media Arts and Technology from UCSB and an MFA from the Ewha Womans University. She had been a tenured art professor at the University of Akron for nine years and is currently a Visiting Professor with emphasis on Art and Machine Learning at the School of Computer Science, Carnegie Mellon University. 

Lingyao Zhang earned her Master's degree in the Machine Learning Department at Carnegie Mellon University.

Manzil Zaheer earned his Ph.D. degree in Machine Learning from the School of Computer Science at Carnegie Mellon University under the able guidance of Prof Barnabas Poczos, Prof Ruslan Salakhutdinov, and Prof Alexander Smola. He is the winner of Oracle Fellowship in 2015. His research interests broadly lie in representation learning. He is interested in developing large-scale inference algorithms for representation learning, both discrete ones using graphical models and continuous with deep networks, for all kinds of data. He enjoys learning and implementing complicated statistical inference, data-parallelism, and algorithms in a simple way.

Dr. Barnabás Póczos is an associate professor in the Machine Learning Department at the School of Computer Science, Carnegie Mellon University. His research interests lie in the theoretical questions of statistics and their applications to machine learning. Currently he is developing machine learning methods for advancing automated discovery and efficient data processing in applied sciences including health-sciences, neuroscience, bioinformatics, cosmology, agriculture, robotics, civil engineering, and material sciences. His results have been published in top machine learning journals and conference proceedings, and he is the co-author of 100+ peer reviewed papers. He has been a PI or co-Investigator on 15+ federal and non-federal grants. Dr. Poczos is a member of the Auton Lab in the School of Computer Science. He is a recipient of the Yahoo! ACE award. In 2001 he earned his M.Sc. in applied mathematics at Eotvos Lorand University in Budapest, Hungary. In 2007 he obtained his Ph.D. in computer science from the same university. From 2007-2010 he was a postdoctoral fellow in the RLAI group at University of Alberta, then he moved to Pittsburgh where he was a postdoctoral fellow in the Auton Lab at Carnegie Mellon from 2010-2012.

\end{document}